\documentclass{amsart}

\makeatletter
\def\enddoc@text{\ifx\@empty\@translators \else\par\bigskip\@settranslators\fi
  \ifx\@empty\addresses \else\null\par\vfill\@setaddresses\fi%
}
\def\@setaddresses{\par
  \nobreak \begingroup
\footnotesize
  \def\author##1{\nobreak\addvspace\bigskipamount}%
  \def\\{\unskip, \ignorespaces}%
  \interlinepenalty\@M
  \def\address##1##2{\begingroup
     \par\addvspace\bigskipamount\noindent
     \@ifnotempty{##1}{{\itshape\ignorespaces##1\unskip}\/: }%
     {\scshape\ignorespaces##2}\par\endgroup}%
  \def\curraddr##1##2{\begingroup
     \@ifnotempty{##2}{\nobreak\noindent{\itshape Current address}%
       \@ifnotempty{##1}{ (\ignorespaces##1\unskip)}\/:\space
       ##2\par\endgroup}}%
  \def\email##1##2{\begingroup
     \@ifnotempty{##2}{\nobreak\noindent{\itshape E-mail address}%
       \@ifnotempty{##1}{ (\ignorespaces##1\unskip)}\/:\space
       \ttfamily##2\par\endgroup}}%
  \def\urladdr##1##2{\begingroup
     \@ifnotempty{##2}{\nobreak\indent{\itshape URL}%
       \@ifnotempty{##1}{ (\ignorespaces##1\unskip)}\/:\space
       \ttfamily##2\par\endgroup}}%
  \addresses
  \endgroup
}
\def\specialsection{\@startsection{section}{1}%
   \z@{\linespacing\@plus\linespacing}{.5\linespacing}%
   {\normalsize\itshape\centering}}
\def\section{\@startsection{section}{1}%
   \z@{\linespacing\@plus\linespacing}{.5\linespacing}%
   {\normalfont\scshape\centering}}
\def\subsection{\@startsection{subsection}{2}%
   \z@{.7\linespacing\@plus.5\linespacing}{-.7em}%
   {\normalfont\bfseries}}
\def\subsubsection{\@startsection{subsubsection}{3}%
   \z@{.5\linespacing\@plus.5\linespacing}{-.7em}%
   {\normalfont\itshape}}
\def\paragraph{\@startsection{paragraph}{4}%
   \z@{.5\linespacing\@plus.3\linespacing}{-.5em}%
   {\normalfont\itshape}}
\def\subparagraph{\@startsection{subparagraph}{5}%
   \z@{.5\linespacing\@plus.3\linespacing}{-.5em}%
   {\normalfont\itshape}}
\def\th@plain{%
  \thm@headfont{\scshape}
  \let\thmhead\thmhead@plain \let\swappedhead\swappedhead@plain
  \thm@preskip.5\baselineskip\@plus.2\baselineskip
                                    \@minus.2\baselineskip
  \thm@postskip\thm@preskip
  \itshape
}
\def\th@definition{%
  \thm@headfont{\bfseries}
  \let\thmhead\thmhead@plain \let\swappedhead\swappedhead@plain
  \thm@preskip.5\baselineskip\@plus.2\baselineskip
                                    \@minus.2\baselineskip
  \thm@postskip\thm@preskip
  \upshape
}
\def\th@remark{%
  \thm@headfont{\itshape}
  \let\thmhead\thmhead@plain \let\swappedhead\swappedhead@plain
  \thm@preskip.5\baselineskip\@plus.2\baselineskip
                                    \@minus.2\baselineskip
  \thm@postskip\thm@preskip
  \upshape
}
\renewenvironment{proof}[1][\proofname]{\par
  \normalfont
  \topsep6\p@\@plus6\p@ \trivlist
  \item[\hskip\labelsep\itshape
    #1\@addpunct{.}]\ignorespaces
}{%
  \qed\endtrivlist
}
\makeatother
\calclayout

\usepackage{natbib}
\usepackage{tree-dvips}
\usepackage{amssymb}
\usepackage{amscd}
\usepackage{latexsym}

\newcommand{\A}{\ensuremath{\mathfrak{A}}}
\newcommand{\B}{\ensuremath{\mathfrak{B}}}
\newcommand{\Lang}{\ensuremath{\mathcal{L}}}
\newcommand{\LIFT}{\ensuremath{\textsf{LIFT}}}
\newcommand{\MSOL}{\ensuremath{\textsf{MSOL}}}
\newcommand{\M}{\ensuremath{\mathcal{M}}}
\newcommand{\F}{\ensuremath{\mathcal{F}}}
\newcommand{\Nat}{\ensuremath{\mathbb{N}}}
\newcommand{\kon}{\raisebox{.5ex}{\ensuremath{\smallfrown}}}
\newcommand{\traRa}{\ensuremath{\stackrel{\ast}{\Rightarrow}\,}}
\newcommand{\tuple}[2][n]{\ensuremath{#2_1,\dots,#2_{#1}}}
\newcommand{\carrier}[1][A]{\ensuremath{\mathcal{#1}}}
\newcommand{\clone}[2][]{\ensuremath{\textsf{Clo}_{#1}(#2)}}

\let\mc=\multicolumn 
\let\func=\longrightarrow 
\let\ra=\rightarrow 
\let\Ra=\Rightarrow

\newtheorem{theorem}{Theorem}

\theoremstyle{definition}
\newtheorem{definition}{Definition}
\theoremstyle{remark}
\newtheorem{example}{Example}

\bibpunct[$\!$]{(}{)}{,}{a}{}{,}

\begin{document}

\title{On Cloning Context-Freeness}
\author{Uwe M\"onnich}
\address{T\"ubingen University\\
SfS\\
Wilhelmstr.~113\\
D-72074 T\"ubingen}
\email{uwe.moennich@uni-tuebingen.de}

\thanks{To appear in: Hans-Peter Kolb \& Uwe M\"onnich: \emph{The
    Mathematics of Syntactic Structure: Trees and their Logics}}
\def\cpright{Der Autor}
\def\copyrightyearmodC{97}
\def\currentmonth{Juni}
\def\currentissue{114}

\maketitle


\section{Introduction}

A first-order language is called monadic if its non-logical vocabulary
lacks both function symbols and relation symbols of arity greater than
one.  Formulas of this kind of language can be mechanically tested for
validity but its expressive power seems to be too restricted to be of
any interest for linguistics.  Monadic second-order languages on the
other hand extend the language of (full) first-order logic by providing
for quantifiable monadic second-order variables.  These languages
increase the definitional resources of first-order logic substantially
but they still retain many metamathematical properties which make them a
`good source of expressive and manageable theories.'
\cite[]{gure:mona87}

In the paper from which the preceding remark is taken Gurevich gives a
detailed account of two strong decidability techniques which were
developed in connection with monadic second-order theories.  One of
these techniques is based on games and automata while the other employs
generalized products.  The applicability of these methods to certain
second-order theories where quantification is restricted to variables
ranging over subsets of the domain of discourse underlines the fact that
formalizations couched in this linguistic framework are supported by a
manageable underlying logic.  In sharp contrast with the situation
exemplified by monadic first-order logic the decidability of applied
forms of its second-order counterpart detracts nothing from its value as
a flexible expressive means for the presentation of theories which
embody interesting parts of either mathematics or the empirical
sciences.

This is not the proper place to substantiate this last claim as far as
mathematical theories are concerned.  The interested reader is referred
to the paper by Gurevich where she can find a wealth of examples.  The
development which brought linguistics into the picture was initiated by
\cite{roge:stud94} who showed in his dissertation that the majority of
technical assumptions which form the core of the Principles and
Parameter incarnation of transformational grammar are conducive to
transparent formalization in monadic second-order tree logic.  Using
Rabin's second-order logic with multiple successors he was able to give
convincing definitions of such central notions as government, chain and
category and, in addition to these positive results, to make a
significant contribution to the theory-internal debate among students of
the Principles and Parameter model by proving that such notions as
free-indexing would make Rabin's logic undecidable when added to its
vocabulary.  Especially this delimiting consequence constitutes a
definite advantage of Roger's approach over competing attempts at
formalizing the theoretical concepts underlying different versions of
transformational grammar.  The main drawback of these attempts consists
in an excess of expressive power, which makes it extremely difficult to
establish any formal criteria to separate the empirically fruitful from
the spurious constructs in the Principles and Parameter model.

Unfortunately, there remains one difficulty which threatens to obstruct
the project of using monadic second-order logic as the proper framework
for a restrictive linguistic metatheory.  This difficulty has its origin
in the close relationship between recognizable tree sets and weak
monadic second-order logic of multiple successors.  It has been known
since the pioneering work of \cite{doner:70} and
\cite{thatcher&wright:68} that the sets of finite trees which can be
generated by regular tree grammars are identifiable with the collections
of finite trees definable by formulae of the weak version of Rabin's
logic.  Since the corresponding string sets, i.e.\ the concatenation of
the labels that decorate the leaves of trees in either of these
equivalent classes, are exactly identical to the context-free languages
the question of whether monadic second-order logic of multiple
successors provides an adequate formal framework for linguistic purposes
is inseparable from the issue of the proper location of natural
languages on the Chomskyan hierarchy.  Evidence from languages like
Swiss-German \cite[]{huybregts84,shie:evid85} and Bambara \cite[]{culy:thec85}
suggests that natural languages exhibit a certain degree of
context-sensitivity and that a formal system which aims at serving as an
adequate mathematical basis for the working linguist must accommodate a
corresponding amount of structural complexity.

There are serious obstacles, however, to any program devoted to a
strengthening of Rabin's logic.  As has been shown by several people,
even minor additions to the vocabulary of the underlying language lead
to undecidability \cite[cf.][]{laeu:mona87}.  The set of theorems of
monadic tree theory, another designation of the theory of multiple
successors which refers to its intended model, seems to be of so high a
degree of complexity that enriching its language always is in danger of
overshooting the mark.  A natural alternative to an extension of the
basic vocabulary of the language would be a restricted form of
second-order polyadic quantification.  In such a move one would allow
relational variables into the language, together with quantifiers
binding them.  In sharp contrast to the monadic set variables which can
be assigned an arbitrary subset of the universe of discourse as a value,
the range of interpretation of these new variables would have to be
narrowed to special relations over the universe of discourse.  Earlier
work on the monadic theory lets one expect that a careful restriction of
the quantificational realm along the indicated lines could generate the
desired supplementary expressiveness without losing control over the
class of theorems.  Of particular significance for the current issue are
the results by \cite{rabin72} and \cite{sief:ther75} to the effect that
restricting quantification to regular subsets leads to a proper
subsystem of the logic, whereas restricting quantification to recursive
subsets leaves the class of theorems untouched.  The kind of relations
in the polyadic case one will have to consider should perhaps not be
determined with reference to their computational complexity, but rather
with reference to their internal structure.  The recent characterization
of context-free languages as sets of strings which can be defined by
first-order sentences with an initial second-order existential
quantifier provides an instructive example for such a semantic approach
in a related context.  Extending the work of \cite{buec:weak60} and
\cite{elgot61} \cite{laut:logi95} have shown that the model sets
satisfying $\Sigma_1^1$ sentences of dyadic second-order logic of one
successor coincide with the context-free languages if the existential
second-order quantifier ranges over binary relations that display the
dependencies of matching parentheses.  It is an open problem whether the
introduction of similarly restricted dyadic quantifiers into the tree
language with multiple successors would be accompanied by a controlled
increase of the definitional resources and in this way vindicate the
idea of a secondary semantics for the limited amount of
context-sensitivity that is exemplified by natural languages.

We cannot end this short outline of possible ways out of the narrow
confines of context-freeness without mentioning free variable
second-order logic.  This logic is expressed in a language in which
relation and function variables have free, but no bound occurrences.
What this means for our monadic second-order tree language is that set
variables may still be syntactically bound by higher-order quantifiers
but the free relation and function variables get an interpretation which
treats them implicitly as universally quantified.  Again, it appears to
be an open question whether this extension of syntactic resources steers
clear of undecidability.

Even though one of the sketched strengthenings of the underlying tree
logic---extension of the non-logical vocabulary, restricted polyadic
second-order quantification or the introduction of free function and
relation variables---may secure just the right increase in expressive
power, the precarious status of the monadic second-order theory of
multiple successors verging on the undecidable gives little cause for
hope.  We shall therefore propose a hybrid solution to the problem of
how to account for mildly context-sensitive phenomena with the help of
tree logic.  The limited expressive power of this logic in its original
set-up makes it impossible to formulate the solution in a way that would
deal directly with the problematic phenomena, but we can give these
phenomena a slightly different appearance whereby they do become
context-free and as such definable in tree logic.

To be somewhat more specific, we will use an ``algebraicized'' variant
of grammars with macro-like productions \cite[]{fischer68} to present an
analysis of structural dependencies which are beyond the reach of
context-free devices.  Macro grammars constitute a natural extension of
traditional rewrite systems.  Whereas in traditional presentations of
rule systems for abstract language families the emphasis has been on a
first-order substitution process in which auxiliary variables are
replaced by elements of the carrier of the proper
algebra---concatenations of terminal symbols and auxiliary category
variables in the string case---macro grammars lift this process to the
second-order level of operations defined on the elements of the carrier
of the algebra.  Once this second-order substitution process is turned
into a process involving only individual variables we are back on
familiar ground, the only difference being the shift from single-sorted
to many-sorted algebras.

When, two paragraphs back, we spoke about two different representations
of context-sensitive dependencies we had this shift from a functional to
an objectual substitution process in mind.  What makes the
transformation of the higher-order into an objectual substitution
possible is the technical device of a derived or lifted alphabet.  In
informal terms, the formation of a derived alphabet can be explained as
a nominalization process turning functions into objects. As a result,
these latter saturated entities have to rely on an explicit composition
operation that provides the necessary glue for combining the objectual
counterpart of the original functional entity with its arguments.  Since
the objectual counterpart now belongs to the type of saturated
entities, the substitution process on the symbolic level, when applied
to the nominalized expressions resulting from the derivation of the
original alphabet, can then take the familiar form of first-order
substitution.  As should be obvious from the preceding remarks, the
language generated by this first-order substitution procedure is
context-free and as such amenable to logical analysis in the framework
of monadic second-order logic with multiple successors.  There is,
however, a negative aspect of the approach using derived alphabets.
During the translation of the original alphabet into its nominalized
derived counterpart new operators representing functional
composition---and, furthermore, new constants standing for projection
functions which we have suppressed in our informal account---are
introduced that change the configurational shape of the translated
expressions.  This point can be made more vivid in terms of a notational
format that is taken for illustrative purposes from lambda calculus with
parallel abstraction.  Let $\langle x_1, \ldots x_n\rangle$ denote
parallel ($n$-ary) abstraction and $comp(\cdot,\cdot)$ binary
application.  The lifted variant of the term $f(a,b)$ then corresponds
to the following expression $comp(\langle x_1, x_2\rangle f(x_1, x_2),
p(a,b))$, where $p$ represents pair formation in the derived alphabet.
This example has to be taken with a grain of salt because in the
parallel lambda calculus both the redex and its converted form are
supposed to have the same meaning whereas in the context of formal
language theory the simple fact that these expressions are different
complex symbols embodies two competing proposals for syntactic analysis.

In the last section of the paper we show how to solve the tension
between the definitions for linguistic concepts in terms of structural
configurations of the original trees and their counterparts in the
lifted context.  While the definitions that are informed by the original
set-up are extensionally inadequate in the general case---they fail to
refer to the context-sensitive dependencies---their lifted counterparts,
living in an environment of first-order substitution, can be combined
with adequate characterizations of those context-sensitive structures.
This combination is made possible by the closure of call by value tree
languages under deterministic bottom-up tree transducer mappings
\cite[see][]{ES78:IO-OI2}.

We have not attempted to present a defense of macro grammars or of a
theory of structural notions embodied in this particular format.  A
detailed discussion of the (de-)merits of macrogrammatical analyses of a
range of syntactic problems is contained in H.-P.~Kolb's
contribution to the present volume.  It is worth emphasizing that our
application of tree theory to context-sensitive structures is not
intended as a justification for a particular form of syntactic analysis.
This task remains to be done and we would be delighted if others
investigated the structural restrictions that characterize a program of
(derived) syntactic macrotheory.

There are several sources that have influenced the ideas reported here.
Apart from the work on the logical characterization of language classes
that was mentioned above the development in universal algebra that led
to a uniform, signature free treatment of varieties has been our main
inspiration.  From the very beginning of this development it has been
obvious to people working in this field that closed sets of operations
are best presented as categories with finite products.  When this
presentation is retranslated into the language of universal algebra we
are confronted with a signature whose only non-constant operators are
symbols for target tupling and functional composition.  Algebras with
signatures of this type will play a major role in the paper and they
will provide the technical foundation for extending the logical methods
of descriptive complexity theory to context-sensitive phenomena.

The first to see the potential for tree language theory of this type of
lifted signature was \cite{maibaum74}.  He showed in particular how to
map context-free into regular tree productions rules. Unfortunately, a
substantial part of his results are wrong because he mistakenly assumed
that for so called call by name derivations an unrestricted derivation
in a lifted signature would leave the generated language unchanged.
\cite{ES77:IO-OI1,ES78:IO-OI2} point out this mistake and give
fixed-point characterizations for both call by name and call by value
context-free production systems.  We hope that the present paper
complements the denotational semantics for call by name tree languages
by giving an operational analysis of the derivation process both on the
level of the original and on the level of the lifted signature.

\section{Preliminaries}
\label{sec:prelim}

The purpose of this section is to fix notations and to present
definitions for the basic notions related to universal algebra. The key
notion is that of a derived algebra.  As was indicated above, the
realization that derived algebras allow for a different presentation
based on a lifted signature constitutes the main step in the process of
restoring context-freeness.  In this context, we will be involved in a
generalization of formal language concepts.  Many-sorted or
heterogeneous algebras provide the proper formal environment to express
this generalization. For expository purposes we will not give every
definition in its most general form but keep to the single-sorted case
where the reader can easily construct the extension to a situation where
more than one sort matter.

\begin{definition}
  Let $S$ be a set of sorts (categories). A \textbf{many--sorted
    alphabet} $\Sigma$ is an indexed family
$
\langle \Sigma_{w,s} \, | \, w\in S^\ast, s\in S \rangle
$
of disjoint sets. A symbol in $\Sigma_{w,s}$ is called an
\textbf{operator} of \textbf{type} $\langle w,s\rangle$, \textbf{arity}
$w$, \textbf{sort} $s$ and \textbf{rank} $l(w)$. If $w=\varepsilon$ then
$f\in\Sigma_{\varepsilon,s}$ is called a \textbf{constant} of sort $s$.
$l(w)$ denotes the length of $w$.
\end{definition}

Note that a ranked alphabet in the traditional terminology can be
identified with an $S$--sorted alphabet where $S=\{s\}$. The set
$\Sigma_{s^n,s}$ is then the same as $\Sigma_n$.
In the way of explanation, let us print out that each symbol $f$ in
$\Sigma_{w,s}$ represents an operation taking $n$ arguments, the $i$th
argument being of sort $w_{i}$, and yielding an element of sort $s$, where
$w=w_{1}\cdots w_{n}$. Alternative designation for many-sorted alphabets
are \emph{many-sorted signatures} or \emph{many-sorted operator
  domains}. We list some familiar examples of single-sorted signatures
for further reference.

\begin{example}
  \hspace*{2cm}
\begin{enumerate}
  \renewcommand{\labelenumi}{\alph{enumi})}
  \renewcommand{\theenumi}{\alph{enumi}}
\item $\Sigma_0 = \{ \varepsilon \}\cup V$ \hspace*{.5cm} $\Sigma_2 =
  \{\kon\}$\\
  Single--sorted signature of semi--groups, extended by a finite set of
  constants $V$.
\item $\Sigma_0 = \{ \varepsilon \}$ \hspace*{.5cm} $\Sigma_1 = \{a \,
  | \, a\in V  \}$\\
  Single--sorted signature of a \emph{monadic} algebra.
\item $\Sigma_2 = \{ \wedge, \vee \}$\\
  Single--sorted signature of lattices.
\end{enumerate}
\end{example}

As was mentioned above, a full description of the theory of a class of
algebras of the same similarity type is given by the totality of the
derived operations. These operations can be indicated by suitably
constructed terms over the basic operators and a set of variables.

\begin{definition}
  For a many--sorted alphabet $\Sigma$, we denote by $T(\Sigma)$ the
  family
$
\langle T(\Sigma,s) \, | \, s\in S \rangle
$
of \textbf{trees of sort} $s$ over $\Sigma$. $T(\Sigma,s)$ is
inductively defined as follows:
\begin{enumerate}
  \renewcommand{\labelenumi}{(\roman{enumi})}
\item For each sort $s\in S$
  \[
  \Sigma_{\varepsilon,s} \subseteq T(\Sigma,s)
  \]
\item For $n\geq 1$ and $s\in S$, $w\in S^\ast$, if $f\in \Sigma_{w,s}$
  and for $1\leq i\leq n$, $t_i\in T(\Sigma,w_i)$, $l(w) =
  n$,
  \[
  f(t_1,\dots,t_n) \in T(\Sigma,s)
  \]
\end{enumerate}
\end{definition}

\begin{definition}
  For a many--sorted alphabet $\Sigma$ and a family of disjoint sets $Y
  = \langle Y_s \, | \, s\in S \rangle$, the family $T(\Sigma,Y)$ is
  defined to be $T(\Sigma(Y))$, where $\Sigma(Y)$ is the many--sorted
  alphabet with $\Sigma(Y)_{\varepsilon,s} = \Sigma_{\varepsilon,s} \cup
  Y_s \text{ and for } w\not= \varepsilon, \,\, \Sigma(Y)_{w,s} =
  \Sigma_{w,s}$ We call a subset \Lang\ of $T(\Sigma, s)$ a \textbf{tree
    language} over $\Sigma$ (of sort $s$).
\end{definition}

Having described the syntax of the tree terms and having indicated their
intended interpretation, it remains to specify the central notion of an
algebra and to give a precise definition of the way in which the formal
term symbols induce an operation on an algebra.

\begin{definition}
  Suppose that $S$ is a set of sorts and that $\Sigma$ is a many--sorted
  alphabet. A $\Sigma$-\textbf{algebra} \A\ consists of an $S$-indexed
  family of sets $\carrier[A] = \langle A_s \rangle_{s\in S}$ and for
  each operator $\sigma \in \Sigma_{w,s}$, a function $\sigma_\A: A^w
  \func A^s$ where $A^w = A^{w_1}\times\cdots\times A^{w_n}$ and $w =
  w_1\cdots w_n$.
  
  The family $\carrier[A]$ is called the sorted \textbf{carrier} of the
  algebra \A\ and is sometimes written $|\A|$.
\end{definition}

Different algebras, defined over the same operator domain, are related
to each other if there exists a mapping between their carriers that is
compatible with the basic structural operations.

\begin{definition}
  A $\Sigma$-\textbf{homomorphism} of $\Sigma$-algebras $h: \A \func \B$
  is an indexed family of functions $h_s: A_s \func B_s$, $(s \in S)$
  such that for every operator $\sigma$ of type $\langle w,s \rangle$ \[
  h_s(\sigma_\A(\tuple{a})) =
  \sigma_\B(h_{w_1}(a_1),\dots,h_{w_n}(a_n))\] for every $n$-tuple
  $(\tuple{a})\in A^w$.
\end{definition}

The set of sorted trees $T(\Sigma,Y)$ can be made into a
$\Sigma$-algebra by defining the operations in the following way. For
every $\sigma$ in $\Sigma_{w,s}$, for every $(\tuple{t})\in T(\Sigma,Y)^w$
\[ \sigma_{\mathfrak{T}(\Sigma,Y)}(\tuple{t}) = \sigma(\tuple{t}).\] 
If \A\ is a $\Sigma$-algebra, each term $t\in T(\Sigma(Y_w),s)$ induces
a function \[t_\A: A^w \func A^s\] called a \emph{derived operation}.
In the place of $Y$ we have used the set of sorted variables $Y_w := \{
y_{i,w_i} \, | \, 1\leq i\leq l(w) \}$. The meaning of the derived
operation $t_\A$ is defined as follows: for $(\tuple{a}) \in A^w$ \[
t_\A(\tuple{a}) = \hat{a}(t)\] where $\hat{a}:
\mathfrak{T}(\Sigma,Y_w)\func \A$ is the unique homomorphism with
$\hat{a}(y_{i,w_i}) = a_1$.

In intuitive terms, the evaluation of a $\Sigma(Y)$-tree $t$ in a given
$\Sigma$-algebra proceeds as follows. First, one assigns a value
$\hat{a}(y_i)\in A$ to every variable in $Y$. Then the operations of \A\ 
are applied to these elements of $A$ as directed by the structure of
$t$.  There is, though, another description of the action performed by
the derived operations on an algebra \A. According to this conception
the derived operations of an algebra \A\ are the mappings one gets from
the projections $y_\A$ by iterated composition with the primitive
operations $\sigma_\A$ $(\sigma\in\Sigma)$.

Given any $\Sigma$-algebra \A\, we can describe the process of
determining the set of derived operations within the framework of
another algebra that is based on a different signature. In this new
signature the symbols of the original alphabet $\Sigma$ are now treated
as constants, as are the projections. The only operations of non-zero
arity are the composition functions and the functions of target tupling.
By \emph{composition of operations} is meant the construction of an
operation $h$ of type $\langle w,s\rangle$ from given operations $f$ of
type $\langle v,s\rangle$ and $g_i$ of type $\langle w,v_i\rangle$ where
$\langle w,s\rangle,\langle v,s\rangle,\langle w,v_i\rangle \in S^\ast
\times S$. The constructed operation $h$ satisfies the rule
\[ h(a) = f(g_1(a),\dots,g_{k}(a))
\]
where $k=|v|$ and $a\in A^w$.

If the operations of \textsl{target tupling} are among the basic
operations that are denoted by the symbols of the new signature, the
type of composition operations can be simplified to $\langle \langle
v,s\rangle,\langle w,v\rangle\rangle,\langle w,s\rangle$. Take again the
operations $g_i$ with their types as just indicated. By their target
tupling is meant the construction of an operation $h$ of type $\langle
w,v\rangle$ that satisfies the rule
\[ h(a) = (g_1(a),\dots,g_k(a)) \]
where again $a\in A^w$ and the outer parentheses on the right-hand side
indicate the ordered $k$-tuple of values. Having introduced composition
and target tupling and having indicated their intended interpretation
the only missing ingredient that remains for us to introduce, before we
can define the concept behind the title of the paper, in the collection
of projection operations.

The \emph{projection operations} on a Cartesian product $A^w$ are the
trivial operations $\pi^w_{w_i}$ satisfying
\[ \pi^w_{w_i}(a) = a_{w_i} \]

A \emph{closed set of operations} or, more briefly, a \emph{clone} on a
family of non-void sets $A=\langle A_s\rangle$ $(s\in S)$ is a set of
operations on $A$ that contains the projection operations and is closed
under all compositions. The relevance of clones for the purposes of
understanding the hidden structure of an algebra and of providing a
signature free treatment for universal algebra was first realized by
P.~Hall. An alternative proposal under the name of algebraic theories
and their algebras is due to J.~B\'enabou and W.~Lawvere. We have chosen
to follow the example of J.~Gallier and of J.~Engelfriet and
E.~M.~Schmidt in using the style of presentation familiar from standard
universal algebra. Our only departure from this tradition is the
explicit inclusion of the operation of target tupling into an official
definition of the clone of term operations.

\begin{definition}
  The \textbf{clone of term operations} of an $\Sigma$-algebra \A,
  denoted by $\clone{A}$, is the smallest set of operations on the
  carrier $\carrier[A]=\langle A_s\rangle$ that contains the primitive
  operations of \A, the projections and is closed under composition and
  target tupling.  The set of all operations of type $\langle
  w,s\rangle$ in $\clone{A}$ is denoted by $\clone[\langle
  w,s\rangle]{A}$.
\end{definition}

To actually present a clone as an algebra both the set of sorts $S$ and
the underlying alphabet $\Sigma$ have to be adapted to the new
situation. The idea to characterize this new situation with the help of
a derived alphabet was first used by J.~B\'enabou and later rediscovered
by T.~S.~E.~Maibaum.

\begin{definition}\label{def:der-alph}
  Let $S$ be a set of sorts and $\Sigma = \langle \Sigma_{w,s}\rangle$
  be an $S$-sorted alphabet $(\langle w,s\rangle\in S^\ast\times S)$.
  The \textbf{derived $(S^\ast\times S^\ast)$-sorted alphabet of
    $\Sigma$} denoted by $D(\Sigma)$, is defined as follows:
\[ D(\Sigma) = \Sigma \cup \left\{ 
    \begin{array}{l}
      \pi_i^v\\
      (\,\,\,)_{w,v}\\
      S_{v,w,s}
    \end{array}
  \right.
\]
where $w\in S^\ast$, $v\in S^+$ and $s\in S$. Each $\pi_i^v$ is a
projection operator of type $\langle \varepsilon,\langle
v,v_i\rangle\rangle$, each $(\,\,\,)_{w,v}$ is a tupling operator and
each $S_{v,w,s}$ is a substitution or composition operator of types
$\langle \langle w,v_1\rangle\cdots\langle w,v_n\rangle,\langle
w,v\rangle\rangle$, respectively $\langle \langle v,s\rangle,\langle
w,v\rangle,\langle w,s\rangle\rangle$ and each $\sigma$ in
$\Sigma_{w,s}$ becomes a constant operator in the derived alphabet
$D(\Sigma)$ of type $\langle \varepsilon,\langle w,s\rangle\rangle$.
\end{definition}

\begin{example}
  We have seen above that trees form the sorted carrier of the
  $\Sigma$-algebra $\mathfrak{T}(\Sigma,Y)$. What is of fundamental
  importance for the further development is the fact that trees with
  variables can also be seen as a $D(\Sigma)$-algebra. The tree
  substitution algebra $\mathfrak{DT}(Y)$ is a $D(\Sigma)$-algebra whose
  carrier of sort $\langle w,s\rangle$ is the set of trees
  $T(\Sigma,Y_w)_s$, i.e.\ the set of trees of sort $s$ that may contain
  variables in $Y_w$.  In order to alleviate our notation we will denote
  this carrier by $T(w,s)$. Carriers of sort $\langle w,v\rangle$ are
  $v$-tuples of carriers of sort $\langle w,v_i\rangle$ and are denoted
  by $T(w,v)$. Each $\sigma$ in $\Sigma_{w,s}$ is interpreted as the
  tree $\sigma(y_{1,w_1},\dots,y_{n,w_n})$, each $\pi_i^v$ is
  interpreted as $y_{i,v_i}$, each $(\,\,\,)_{w,v}(\tuple{t})$ is
  interpreted as the formation of $v$-tuples $(\tuple{t})$, where each
  $t_i$ is an element of $T(w,v_i)$ and $l(v) = k$, and each $S_{v,w,s}$
  is interpreted as a composition or substitution of trees. An intuitive
  description of the composite
  $S_{v,w,s_{\mathfrak{DT}(\Sigma,Y)}}(t,t')$ with $t\in T(v,s)$ and $t'
  = (\tuple[k]{t})\in T(w,v)$ is this: the composite is the term of type
  $\langle w,s\rangle$ that is obtained from $t$ by substituting the
  term $t_i$ of sort $v_i$ for the variable $y_{i,v_i}$ in $t$. The
  formal definition of the composition operation relies on the unique
  homomorphism $\hat{t}': \mathfrak{T}(\Sigma,Y_v) \func
  \mathfrak{T}(\Sigma,Y_w)$ that extends the function $t': Y_v \func
  \bigcup_i T(w,v_i)$ mapping $y_{i,v_i}$ to $t_i$ in $T(w,v_i)$. Then
  for any $t\in T(v,s)$, we define
  $S_{v,w,s_{\mathfrak{DT}(\Sigma,Y)}}(t,t')$ as the value of $\hat{t}'$
  on the term $t$:
  \[ S_{v,w,s_{\mathfrak{DT}(\Sigma,Y)}}(t,t') := \hat{t}'(t) =
  t[\tuple{t}]\] where the last term indicates the result of
  substituting $t_i$ for $y_{i,v_i}$.
  
  Since the derived alphabet $D(\Sigma)$ leads to a tree algebra
  $\mathfrak{T}(D(\Sigma))$ in the same way that the alphabet $\Sigma$
  led to the algebra $\mathfrak{T}(\Sigma)$, there is a unique
  homomorphism $\beta: \mathfrak{T}(D(\Sigma)) \func
  \mathfrak{DT}(\Sigma)$. It was pointed out by \cite{gall:n-ra84} that
  this homomorphism is very similar to $\beta$-conversion in the
  $\lambda$-calculus. The explicit specification of its action repeats
  in a concise form the description of the tree substitution algebra:
 \[
 \begin{array}{rcll} 
   \beta_{w,s}(\sigma) & = &
   \sigma(y_{1,w_1},\dots,y_{n,w_n}) & \text{
     for } \sigma \in \Sigma_{w,s} \\
   &&&\\
   \beta_{w,s}(\pi_i^w) & = & y_{i,w_i} & \text{ if } w_i = s \\
   &&&\\
   \beta_{w,s}((\,\,\,)_{w,v}(\tuple[k]{t}) & = & (\tuple[k]{t}) &
   \text{ for } t_i\in T(w,v_i) \\
   &&&\\
   \beta_{w,s}(S_{v,w,s}(t,t')) & = &
   \mc{2}{l}{\beta_{v,s}(t)[\beta_{w,v_1}(t_1),\dots, \beta_{w,v_k}(t_k)]}\\
   && \mc{2}{l}{\hspace*{4em}\text{ for } t\in T(v,s)\text{, } t_i \in
     T(w,v_i)}\\
   && \mc{2}{l}{\hspace*{4em}\text{ and } t' = (\tuple[k]{t}).}
  \end{array}
  \]
\end{example}

\begin{example}
  Suppose that the set of sorts $S$ is a singleton and that $\Sigma$
  contains three symbols $f,a$ and $b$ where $\Sigma_{ss,s}= \{ f \}$
  and $\Sigma_{\varepsilon,s} = \{ a,b \}$ and $s$ is the single sort in
  $S$. As is customary in the context of single-sorted alphabets we
  shall write the type $\langle s^n,s\rangle$ as $n$.  According to this
  notational convention the following figure displays a tree $t$ in
  $T(D(\Sigma))$:

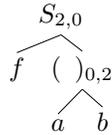
\begin{center}
  \begin{tabular}{ccc}
    \mc{3}{c}{\node{s}{$S_{2,0}$}}\\[2ex]
    \node{f}{$f$} & \mc{2}{c}{\node{br}{$(\,\,\,)_{0,2}$}}\\[2ex]
    & \node{a}{$a$} & \node{b}{$b$} \\
  \end{tabular}
  \nodeconnect[b]{s}[t]{f} \nodeconnect[b]{s}[t]{br}
  \nodeconnect[b]{br}[t]{a} \nodeconnect[b]{br}[t]{b}
\end{center}

Applying $\beta_0$ to this tree it returns as a value the tree
$\beta_0(t) = f(a,b)$ in $T(\Sigma)$.  Displayed in tree form this last
term looks as follows:
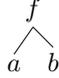
\begin{center}
  \begin{tabular}{cc}
    \mc{2}{c}{\node{f}{$f$}}\\[2ex]
    \node{a}{$a$} & \node{b}{$b$} \\
  \end{tabular}
  \nodeconnect[b]{f}[t]{a} \nodeconnect[b]{f}[t]{b}
\end{center}
\end{example}

The unique homomorphism $\beta$ from the derived tree algebra into the
tree substitution algebra has a right inverse $\LIFT^\Sigma = (
\textsf{LIFT}_{w,s}^\Sigma)(\langle w,s\rangle\in S^\ast \times S)$,
where $\LIFT_{w,s}^\Sigma$ maps a $\Sigma$-tree in $T(\Sigma(Y_w),s)$
into a $D(\Sigma)$-tree in $T(D(\Sigma),\langle \varepsilon,\langle
w,s\rangle\rangle)$.  Since we will have no occasion to apply the
function \LIFT\ to trees over a many-sorted alphabet we content
ourselves with giving its recursive definitions for the single-sorted
case:
                                
\begin{definition}\label{def:lift}
  Suppose that $\Sigma$ is a single-sorted or ranked alphabet. For
  $k\geq 0$, $\LIFT_k^\Sigma: T(\Sigma,X_k) \func T(D(\Sigma),\langle
  \varepsilon,k\rangle)$ is the function defined recursively as follows
  (where $X_k = \{ x_i \, | \, 1 \leq i \leq k \}$):
\begin{eqnarray*}
  \LIFT_k^\Sigma(x_i) & = & \pi_i^k \\
  \LIFT_k^\Sigma(\sigma) & = & S_{0,k,1}(\sigma) \text{ for } \sigma \in
  \Sigma_0\\
  \LIFT_k^\Sigma(\sigma(\tuple{t})) & = & S_{n,k,1}(\sigma,
  (\,\,\,)_{k,n}, (\LIFT_k^\Sigma(t_1),\dots,\LIFT_k^\Sigma(t_n))
  \\
  & & \hspace*{1em} \text{ for } \sigma \in \Sigma_n.
\end{eqnarray*}
\end{definition}

It should be obvious that for any tree $t$ in $T(\Sigma,X_k)$ \[
\beta_k(\LIFT_k^\Sigma(t)) = t.\]

The reader will have noticed that the tupling operator was conspicuous
by its absence in the recursive definition of the \LIFT-function.
According to our official definition for the carrier of the tree
substitution algebra a term like $\sigma(\tuple{t})$ is the result of
composing the $n$-tuple $\tuple{t}$ of $k$-ary terms $t_i$ with the
$n$-ary term $\sigma(\tuple{x})$. This part of the ``structural
history'' is suppressed in the third clause of the \LIFT-specification.
We shall adhere to this policy of eliminating this one layer in the
configurational set-up of tree terms and we shall extend this policy to
the level of explicit trees over the derived alphabet $D(\Sigma)$. This
does not mean that we revoke the official, pedantic definition of the
symbol set in a derived alphabet, but we shall make our strategy of
notational alleviation type consistent by reading the type of each
substitution operator $S_{v,w,s}$ as $\langle \langle v,s\rangle\langle
w,v_1\rangle\cdots\langle w,v_n\rangle,\langle w,s\rangle\rangle$
instead of $\langle \langle v,s\rangle\langle w,v\rangle,\langle
w,s\rangle\rangle$.

\section{Context-Free Tree Languages}

The correspondence between trees in explicit form, displaying
composition and projection labels, and their converted images as
elements of a tree substitution algebra is an example of a situation
which is characterized by a meaning preserving relationship between two
algebras \A\ and \B. Of particular interest to formal language theory is
the situation where problems in an algebra \B\ can be lifted to the tree
level, solved there and, taking advantage of the fact that trees can be
regarded as denoting elements in an arbitrary algebra, projected back
into their original habitat \B. This transfer of problems to the
symbolic level would, of course, produce only notational variants as
long as the lifted environment constitutes just an isomorphic copy of
the domain in relation to which the problems were first formulated. One
might suspect that $\beta$-conversion and its right inverse are a case
of this type. Despite their suspicious similarity, trees in explicit
form and their cousins which are the results of performing the
operations according to the instructions suggested by the composition
and projection symbols, are sufficiently different to make the transfer
of problems to the explicit variants a worthwhile exercise.

In intuitive terms, the difference between the two tree algebras is
related to the difference between a first-order and a second-order
substitution process in production systems. Let us view grammars as a
mechanism in which local transformations on trees can be performed in a
precise way. The central ingredient of a grammar is a finite set of
productions, where each production is a pair of trees. Such a set of
productions determines a binary relation on trees such that two trees
$t$ and $t'$ stand in that relation if $t'$ is the result of removing in
$t$ an occurrence of a first component in a production pair and
replacing it by the second component of the same pair. The simplest type
of such a replacement is defined by a production that specifies the
substitution of a single-node tree $t_0$ by another tree $t_{1}$. Two
trees $t$ and $t'$ satisfy the relation determined by this simple
production if the tree $t'$ differs from the tree $t$ in having a
subtree $t_{1}$ that is rooted at an occurrence of a leaf node $t_{0}$
in $t$. In slightly different terminology, productions of this kind
incorporate instructions to rewrite auxiliary variables as a complex
symbol that, autonomously, stands for an element of a tree algebra. As
long as the carrier of a tree algebra is made of constant tree terms the
process whereby nullary variables are replaced by trees is analogous to
what happens in string languages when a nonterminal auxiliary symbol is
rewritten as a string of terminal and non-terminal symbols,
independently of the context in which it occurs. The situation changes
dramatically if the carrier of the algebra is made of symbolic
counterparts of derived operations and the variables in production rules
range over such second-level entities. As we have seen in the preceding
sections, the tree substitution algebra provides an example for an
algebra with this structure. The following example illustrates the gain
in generative power to be expected from production systems determining
relations among trees that derive from second-order substitution of
operators rather than constants.

\begin{example}
  Let $V$ be a finite vocabulary. It gives rise to a \textsl{monadic}
  signature $\Sigma$ if all the members of $V$ are assigned rank one and
  a new symbol $\varepsilon$ is added as the single constant of rank
  zero. For concreteness, let us assume that we are dealing with a
  vocabulary $V$ that contains the symbols $a$ and $b$ as its only
  members. Trees over the associated monadic signature $\Sigma=\Sigma_0
  \cup \Sigma_1$ where $\Sigma_{0}=\{\varepsilon\}$ and
  $\Sigma_{1}=\{a,b\}$ are arbitrary sequences of applications of the
  operators $a$ and $b$ to the constant $\varepsilon$. It is well known,
  as was pointed out above, that there is a unique homomorphism from
  these trees, considered as the carrier of a $\Sigma$-algebra to any
  other algebra of the same similarity type. In particular, there is a
  homomorphism into $V^\ast$ when $a$ and $b$ are interpreted as
  left-concatenation with the symbol $a$ and $b$, respectively, and when
  $\varepsilon$ is interpreted as the constant string of length zero.
  This homomorphism establishes a bijection between $T(\Sigma)$ and
  $V^\ast$ \cite[cf.][]{maibaum74}. When combined with this bijective
  correspondence the following regular grammar generates the set of all
  finite strings over $V$.

\[
G = \langle \Sigma, \F, S, P \rangle
\]
\[
\begin{array}{lclclcl}
  \Sigma_0 & = & \{\varepsilon\} & \hspace*{1cm} & \Sigma_1 &
  = &  \{a,b\}\\
  \F_0 & = & \{S\} & \hspace*{1cm} & \F_n & = & \emptyset \mbox{ for } n
  \geq 1
\end{array}
\]
\[
P = \{ S\ra \varepsilon\, | \, a(S)\, | \, b(S) \}
\]
\[
\mathcal{L}(G,S) = \Sigma_1^\ast
\]
where we have identified $T(\Sigma)$ with $\Sigma_1^\ast$.

$P$ stands for the finite set of productions and $s$ stands for the only
non-terminal symbol. $V$ gives also rise to a \textsl{binary} signature
$\Sigma'$ if the members of $V$ are assigned rank zero and two new
symbols are added, $\varepsilon$ of rank zero and $\kon$ of rank two.
Trees over this signature are nonassociative arc-links between the
symbols ${a}$ and ${b}$. When ${a}$ and ${b}$ are interpreted as
constant strings of length one, $\varepsilon$ as constant string of
length zero and the arc $\kon$ as (associative) concatenation ${V^\ast}$
becomes an $\Sigma'$-algebra. Note that the unique homomorphism from
$\mathfrak{T}(\Sigma')$ to $V^\ast$ is \emph{not} a bijection this time.
When combined with this homomorphism the following grammar generates the
string language $\{ a^nb^n\}$.

\[
G' = \langle \Sigma', \F, S, P' \rangle
\]
\[
\begin{array}{lclclcl}
  \Sigma_0' & = & \{\varepsilon, a,b\} & \hspace*{1cm} & \Sigma_2' &
  = &  \{\kon\}\\
  \F_0 & = & \{S\} & \hspace*{1cm} &
\end{array}
\]
\[
P' = \{ S\ra \varepsilon\, | \, \kon(a,\kon(S,b)) \}
\]
\[
\mathcal{L}(G',S) = \{ \kon(a,\kon(\dots,\kon(\varepsilon,b)\dots
b)\dots)\}
\]
where $n$ occurrences of $a$ precede the same number of occurrences of
$b$ for $n\geq 0$.

Now, consider the $D(\Sigma)$-tree substitution algebra
$\mathfrak{DT}(\Sigma,X)$. As will be recalled, its carrier consists of
trees in $T(\Sigma,X)$. Due to the fact that $\Sigma$ is a
\emph{monadic} signature trees in $T(\Sigma,X)$ may not contain more
that a single variable. Auxiliary symbols ranging over this carrier take
therefore as values monadic or constant derived operators. When monadic
auxiliary symbols appear in productions this means that they behave the
way that nullary auxiliary symbols do except for the fact that their
argument has to be inserted into the unique variable slot of their
replacing derived operator. After these explanations it should be
obvious that the following grammar over the original \emph{monadic}
signature $\Sigma$ generates the context-free string language $\{
a^nb^n\}$ when combined with the unique homomorphism mentioned above:

\[
G'' = \langle \Sigma, \F'', S, P'' \rangle
\]
\[
\begin{array}{lclclcl}
  \F_0'' & = & \{s\} & \hspace*{1cm} & \F_1'' & = & \{ F\}
\end{array}
\]
\[
P'' = \left\{ \begin{array}{lcl}
    S & \!\ra\! & F(\varepsilon)\, |\,\varepsilon  ,\\
    F(x) &\! \ra\! & a(F(b(x))) \, | \, a(b(x))
\end{array}\right\}
\]
\[
\mathcal{L}(G'',S) = \{
\overbrace{a(a\dots}^n\overbrace{(b(b\dots}^n(\varepsilon)\dots)\}
\]

\end{example}

The last grammar in the preceding example illustrates for the language
$\{ a^nb^n\}$ a transformation that can be applied to the grammar of any
given context-free language. Employing terminology to be introduced in a
moment, we have the precise characterization of the gain in generative
capacity resulting from the introduction of monadic operator variables
into production systems: For every finite alphabet $V$, the
context-free subsets of $T(\Sigma_V)$ (where $\Sigma_{V}$ is the monadic
signature induced by $V$) have exactly the context-free string languages
as values of the unique homomorphism that maps trees in $T(\Sigma_V)$
into strings in $V^\ast$ and interprets the elements of $V$ as unary
operations that concatenate for the left with their argument.

We now turn to introducing the notion of a context-free tree grammar.
This type of grammar is related to the type of grammars that were defined
by \cite{fischer68} and were called macro grammars by him. Context-free
tree grammars constitute an algebraic generalization of macro grammars
since the use of macro-like productions served the purpose of making
simultaneous string copying a primitive operation.

\begin{definition}
  A \textbf{context--free tree grammar} $G = \langle \Sigma, \F, S, P
  \rangle$ is a 4-tuple, where $\Sigma$ is a finite ranked alphabet of
  \emph{terminals}, $\F$ is a finite ranked alphabet of
  \emph{nonterminals} disjoint from $\Sigma$, $S\in \F$ is the start
  symbol of rank $0$ and $P$ is a finite set of rules of the form
\[
F(x_1,\dots,x_m) \ra t \hspace*{1cm} (n \geq 0)
\]
where $F\in \F_m$ and $t\in T(\Sigma\cup\F,X_m)$. Recall that $X$ is
here assumed to be a set of variables $X= \{x_1,x_2,\dots\}$ and $X_m =
\{x_1,\dots,x_m\}$.
\end{definition}

For reasons having to do with the impossibility of mirroring the process
of copying in a grammar with a completely uncontrolled derivation regime
we restrict ourselves to one particular mode of derivation. According to
this mode a function symbol may be replaced only if all its arguments
are trees over the terminal alphabet. In the conventional case this form
of replacement mechanism would correspond to a rightmost derivation. In
terms of modes of computations, the exclusion of nonterminals that have
any nonterminals appearing in any of their arguments corresponds to a
\textsl{call by value} computation where the actual parameters of a
function call have to be values from the domain of computation.

\begin{definition}
  Let $G = \langle \Sigma, \F, S, P \rangle$ be a context-free tree
  grammar and let $t,t' \in T(\Sigma\cup\F)$.  $t'$ is
  \textbf{directly derivable} by an \textbf{inside-out step} from $t$ ($
  t \Ra t' $) if there is a tree $t_0 \in T(\Sigma\cup\F,X_{1})$
  containing exactly \emph{one} occurrence of $x_{1}$, a corresponding
  rule $F(x_1,\dots,x_m) \ra t''$, and trees $t_1,\dots,t_m \in
  T(\Sigma)$ such that
\begin{eqnarray*}
  t & = & t_0[F(t_1,\dots,t_m)]\\
  t' & = & t_0[t''[t_1,\dots,t_m]]
\end{eqnarray*}
$t'$ is obtained from $t$ by replacing an occurrence of a subtree
$F(t_1,\dots,t_m)$ by the tree $t''[t_1,\dots,t_m]$. By the inside-out
restriction on the derivation scheme it is required that the trees
$t_1,t_2$ through $t_n$ be terminal trees.

Recall from the preceding section that for $m,n\geq 0$, $t\in
T(\Sigma,X_m)$ and $t_1,\dots,t_m \in T(\Sigma,X_n)$ $ t[t_1,\dots,t_m]
$ denotes the result of \textbf{substituting} $t_i$ for $x_i$ in $t$.
Observe that $t[t_1,\dots,t_m]$ is in $T(\Sigma,X_n)$.

As is customary \traRa\ denotes the transitive-reflexive closure of
$\Ra$.
\end{definition}

\begin{definition}
  Suppose $G = \langle \Sigma, \F, S, P \rangle$ is a context-free tree
  grammar. We call
\[
\mathcal{L}(G,S) = \{ t\in T(\Sigma) \, | \, S \traRa t \}
\]
the \textbf{context-free inside-out tree language} generated by $G$ from
$S$.
\end{definition}

We reserve a special definition for the case where $\F$ contains only
function symbols of rank zero.

\begin{definition}
  A \textbf{regular tree grammar} is a tuple $G = \langle \Sigma, \F, S,
  P \rangle$, where $\Sigma$ is a finite ranked alphabet of terminals,
  $\F$ is a finite alphabet of function or nonterminal symbols of rank
  zero, $S\in\F$ is the start symbol and $P\subseteq \F\times
  T(\Sigma\cup\F)$ is a finite set of productions.  The \textbf{regular
    tree language} generated by $G$ is
\[ \Lang = \{ t\in T(\Sigma)\, | \, S \traRa t \}
\]
\end{definition}

Note that in the case of regular grammars the analogy with the
conventional string theory goes through. There is an equivalence of the
unrestricted, the rightmost and the leftmost derivation modes where the
terms 'rightmost' and 'leftmost' are to be understood with respect to
the linear order of the leaves forming the frontier of a tree in a
derivation step.

Very early in the development of (regular) tree grammars it was realized
that there exists a close relationship between the families of trees
generated by tree grammars and the family of context-free string
languages. This fundamental fact is best described by looking at it from
the perspective on trees that views them as symbolic representations of
values in arbitrary domains.  Recall the unique homomorphism from the
introductory example of this section that mapped non-associative
concatenation terms into strings of their nullary constituents. This
homomorphism is a particular case of a mapping than can easily be
specified for an arbitrary signature.

\begin{definition}
  Suppose $\Sigma$ is multi-sorted or ranked alphabet. We call
  \textbf{yield} or \textbf{frontier} the unique homomorphism $y$ that
  interprets every operator in $\Sigma_{w}$ or $\Sigma_{n}$ with $l(w)
  =n$ as the $n$-ary operation of concatenation. More precisely
\[
\begin{array}{rcll}
  y(\sigma) & = & \sigma & \text{ for } \sigma \in \Sigma_\varepsilon
  \text{ (or } \Sigma_0\text{)}\\
  y(\sigma(\tuple{t})) & = & y(t_1)\dots y(t_n) & \text{ for } \sigma
  \in \Sigma_w
  \text{ (or } \Sigma_n\text{) and }\\
  &&& \text{ } t_i\in T(\Sigma)_{w_i} \text{ (or } T(\Sigma)\text{)}
\end{array}
\]
\end{definition}

\smallskip\noindent\textbf{Fact} \textit{
  A (string) language is context-free iff it is the yield of a regular
  tree language.}
\smallskip

As was shown in the introductory example, the addition of macro operator
variables increases the generative power of context-free tree grammars
over monadic alphabets considerably. The following example demonstrates
that the addition of $n$-ary macro operator variables leads to a
significant extension with respect to arbitrary ranked alphabets. The
string language of the following context-free tree language is not
context-free.

\begin{example} 
  Let us consider a context-free tree grammar $G = \langle \Sigma, \F,
  S, P \rangle$ such that its frontier is the set of all
  cross-dependencies between the symbols $a$,$c$ and $b$, $d$,
  respectively. The grammar $G$ consists of the components as shown
  below:

\[
\begin{array}{lclclcl}
  \Sigma_0 & = & \{\varepsilon, a,b,c,d\} & \hspace*{1cm} & \Sigma_2 &
  = &  \{\kon\}\\
  \F_0 & = & \{S\} & \hspace*{1cm} & \F_4 & = & \{ F\}
\end{array}
\]
\[
P = \left\{ \begin{array}{rcl} S& \!\ra\!
    &F(a,\varepsilon,c,\varepsilon)\, |\,
    F(\varepsilon,b,\varepsilon,d)\, | \, \varepsilon ,\\
    F(x_1,x_2,x_3,x_4)&\! \ra\!& \\
    \mc{3}{l}{\hspace*{2cm}F(\kon(a,x_1),x_2,\kon(c,x_3),x_4)\, |}\\
    \mc{3}{l}{\hspace*{2cm}F(x_1,\kon(b,x_2),x_3,\kon(d,x_4))\, |}\\
    \mc{3}{l}{\hspace*{2cm}\kon(\kon(\kon(x_1,x_2),x_3),x_4)} \\
\end{array}\right\} 
\]
\begin{align*}
  \mathcal{L}&(G,S) =\\
  &\{\kon(\kon(\kon(\kon(a,\dots)\dots),\kon(b,\dots)\dots),\kon(c,\dots)\dots),\kon(d,\dots)\dots))\}
\end{align*}

The number of occurrences of $a$'s and $c$'s and of $b$'s and $d$'s,
respectively, has to be the same. By taking the frontier of the tree
terms, we get the language $\mathcal{L}'=\{a^nb^mc^nd^m\}$.
\end{example}

The language of the preceding example illustrates a structure that can
actually be shown to exist in natural language. Take the following
sentences which we have taken from Shieber's (1985) paper:
\begin{example}
  \hspace*{2cm}
\begin{enumerate}
  \renewcommand{\labelenumi}{(\roman{enumi})}
  \renewcommand{\labelenumii}{\alph{enumii})}
  \renewcommand{\theenumi}{\roman{enumi}}
  \renewcommand{\theenumii}{\alph{enumii}}
\item
\begin{enumerate}
\item \emph{Jan s\"ait das mer em Hans es huus h\"alfed aastriiche.}
\item John said that we helped Hans (to) paint the house.
\end{enumerate}
\item
\begin{enumerate}
\item \emph{Jan s\"ait das mer d'chind em Hans es huus l\"ond h\"alfed
    aastriiche.}
\item John said that we let the children help Hans paint the house.
\end{enumerate}
\end{enumerate}
\end{example}

The NP's and the V's of which the NP's are objects occur in cross-serial
order. \emph{D'chind} is the object of \emph{l\"ond}, \emph{em Hans} is
the object of \emph{h\"alfe}, and \emph{es huus} is the object of
\emph{aastriiche}.  Furthermore the verbs mark their objects for case:
\emph{h\"alfe} requires dative case, while \emph{l\"ond} and
\emph{aastriiche} require the accusative. It appears that there are no
limits on the length of such constructions in grammatical sentences of
Swiss German. This fact alone would not suffice to prove that Swiss
German is not a context-free string language. It could still be the case
that Swiss German \emph{in toto} is context-free even though it subsumes
an isolable context-sensitive fragment. Relying on the closure of
context-free languages under intersection with regular languages
\cite{huybregts84} and \cite{shie:evid85}
are able to show that not only the fragment exhibiting the
cross-dependencies but the whole of Swiss German has to be assumed as
non context-free.

Shieber intersects Swiss German with the regular language given in
Example \ref{swiss2} in (\ref{swiis:reg1}) to obtain the result in
(\ref{swiis:res}). As is well known, this language is not context-free.

\begin{example}\label{swiss2}
  \hspace*{2cm}
\begin{enumerate}
  \addtocounter{enumi}{2} \renewcommand{\labelenumi}{(\roman{enumi})}
  \renewcommand{\labelenumii}{\alph{enumii})}
  \renewcommand{\theenumi}{\roman{enumi}}
  \renewcommand{\theenumii}{\alph{enumii}}
\item
\begin{enumerate}
\item \emph{Jan s\"ait das mer (d'chind)$^\ast$ (em Hans)$^\ast$ h\"and
    wele (laa)$^\ast$ (h\"alfe)$^\ast$ aastriiche.}\label{swiis:reg1}
\item John said that we (the children)$^\ast$ (Hans)$^\ast$ the house
  wanted to (let)$^\ast$ (help)$^\ast$ paint.\label{swiis:reg2}
\end{enumerate}
\item \emph{Jan s\"ait das mer (d'chind)$^n$ (em Hans)$^m$ h\"and wele
    (laa)$^n$ (h\"alfe)$^m$ aastriiche.}\label{swiis:res}
\end{enumerate}
\end{example}

Swiss German is not an isolated case that one could try to sidestep and
to classify as methodologically insignificant. During the last 15 years
a core of structural phenomena has been found in genetically and
typologically unrelated languages that leaves no alternative to
reverting to grammatical formalisms whose generative power exceeds that
of context-free grammars.

It has to be admitted that the use of macro-like productions is not the
only device that has been employed for the purpose of providing grammar
formalisms with a controlled increase of generative capacity.
Alternative systems that were developed for the same purpose are e.g.
tree adjoining grammars, head grammars and linear indexed grammars.
Although these systems make highly restrictive claims about natural
language structure their predictive power is closely tied to the
individual strategy they exploit to extend the context-free paradigm.
The great advantage of the tree oriented formalism derives from its
connection with \emph{descriptive complexity theory}. Tree properties
can be classified according to the complexity of logical formulas
expressing them. This leads to the most perspicuous and fully
\emph{grammar independent} characterization of tree families by monadic
second-order logic.  Although this characterization encompasses only
regular tree sets the lifting process of the preceding section allows us
to simulate the effect of macro-like productions with regular rewrite
rules.

Again, the device of lifting an alphabet into its derived form is not
without its alternatives in terms of which a regular tree set can be
created that has as value the intended set of tree structures over the
original alphabet. Our own reason for resting with the lifting process
was the need to carry through the "regularizing" interpretation not only
for the generated language, but also for the derivation steps.

A very simple example is now given of a context-free tree grammar that
specifies as its frontier the (non-context-free) string language
$\{a^nb^nc^n\}$. At the same time we shall present the lifted version of
the grammar and illustrate the effect of the productions with two
sequences of derivation trees.

\begin{example} 
  Consider the context-free tree grammar $G = \langle \Sigma, \F,
  S, P \rangle$ which consists of the components as shown below:
\[
\begin{array}{lclclcl}
  \Sigma_0 & = & \{a,b,c\} & \hspace*{1cm} & \Sigma_2 &
  = &  \{\kon\}\\
  \F_0 & = & \{S\} & \hspace*{1cm} & \F_3 & = & \{ F\}
\end{array}
\]
\[
P = \left\{ \begin{array}{rcl}
    S& \!\ra\! &F(a,b,c)\, \\
    F(x_1,x_2,x_3)&\! \ra\!& \\
    \mc{3}{l}{\hspace*{2cm}F(\kon(a,x_1),\kon(b,x_2),\kon(c,x_3))\, |}\\
    \mc{3}{l}{\hspace*{2cm}\kon(\kon(x_1,x_2),x_3))} \\
\end{array}\right\}
\]

Applying the $S$-production once and the first $F$-production two times
we arrive at the sequence of trees in Figure \ref{fig:anbncnfirst}.

\begin{figure}[htp]
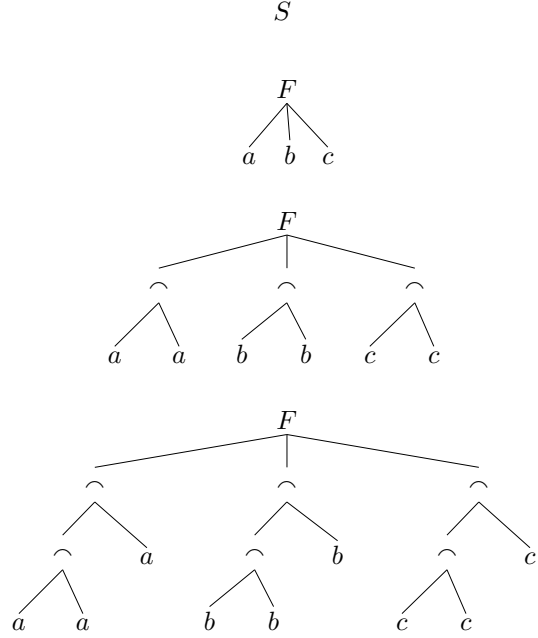

  \begin{center}
    \leavevmode $S$ \vspace*{2em}

    \begin{tabular}{ccc}
      \mc{3}{c}{\node{F}{$F$}}\\[3ex]
      \node{a}{$a$} &      \node{b}{$b$} &      \node{c}{$c$}\\
      \nodeconnect[b]{F}[t]{a} \nodeconnect[b]{F}[t]{b}
      \nodeconnect[b]{F}[t]{c}
    \end{tabular}

    \begin{tabular}{p{.5cm}p{.5cm}p{.5cm}p{.5cm}p{.5cm}p{.5cm}}
      \mc{6}{c}{\node{F}{$F$}}\\[3ex]
      \mc{2}{c}{\node{k1}{$\kon$}} & \mc{2}{c}{\node{k2}{$\kon$}} &
      \mc{2}{c}{\node{k3}{$\kon$}}\\[3ex]
      \node{a1}{$a$} & \node{a2}{$a$} & \node{b1}{$b$} & \node{b2}{$b$}
      &
      \node{c1}{$c$} & \node{c2}{$c$}\\
      \nodeconnect[b]{F}[t]{k1} \nodeconnect[b]{F}[t]{k2}
      \nodeconnect[b]{F}[t]{k3} \nodeconnect[b]{k1}[t]{a1}
      \nodeconnect[b]{k1}[t]{a2} \nodeconnect[b]{k2}[t]{b1}
      \nodeconnect[b]{k2}[t]{b2} \nodeconnect[b]{k3}[t]{c1}
      \nodeconnect[b]{k3}[t]{c2}
    \end{tabular}

    \begin{tabular}{p{.5cm}p{.5cm}p{.5cm}p{.5cm}p{.5cm}p{.5cm}p{.5cm}p{.5cm}p{.5cm}} 
      \mc{9}{c}{\node{F}{$F$}}\\[3ex]
      \mc{3}{c}{\node{k1}{$\kon$}} & \mc{3}{c}{\node{k2}{$\kon$}} &
      \mc{3}{c}{\node{k3}{$\kon$}}\\[3ex]
      \mc{2}{c}{\node{k4}{$\kon$}} & \node{a3}{$a$} &
      \mc{2}{c}{\node{k5}{$\kon$}} & \node{b3}{$b$} &
      \mc{2}{c}{\node{k6}{$\kon$}} & \node{c3}{$c$}\\[3ex]
      \node{a1}{$a$} & \node{a2}{$a$} && \node{b1}{$b$} & \node{b2}{$b$}
      &&
      \node{c1}{$c$} & \node{c2}{$c$} &\\
      \nodeconnect[b]{F}[t]{k1} \nodeconnect[b]{F}[t]{k2}
      \nodeconnect[b]{F}[t]{k3} \nodeconnect[b]{k1}[t]{k4}
      \nodeconnect[b]{k1}[t]{a3} \nodeconnect[b]{k2}[t]{k5}
      \nodeconnect[b]{k2}[t]{b3} \nodeconnect[b]{k3}[t]{k6}
      \nodeconnect[b]{k3}[t]{c3} \nodeconnect[b]{k4}[t]{a1}
      \nodeconnect[b]{k4}[t]{a2} \nodeconnect[b]{k5}[t]{b1}
      \nodeconnect[b]{k5}[t]{b2} \nodeconnect[b]{k6}[t]{c1}
      \nodeconnect[b]{k6}[t]{c2}
    \end{tabular}

    \caption{An Example for $\{a^nb^nc^n\}$}
    \label{fig:anbncnfirst}
  \end{center}
\end{figure}

The result of applying the terminal $F$-production to the last three
trees in Figure \ref{fig:anbncnfirst} is shown in Figure
\ref{fig:anbncnsecond}.

\begin{figure}[htp]
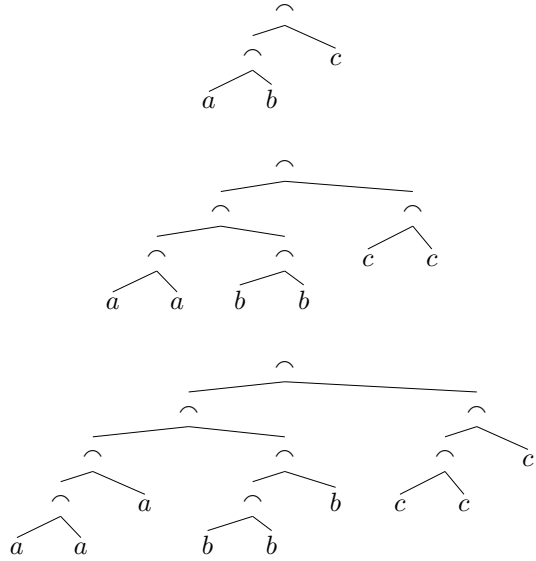

  \begin{center}
    \leavevmode
      \begin{tabular}{p{.5cm}p{.5cm}p{.5cm}}
        \mc{3}{c}{\node{k1}{$\kon$}}\\[1ex]
        \mc{2}{c}{\node{k2}{$\kon$}} & \node{c}{$c$}\\[1ex]
        \node{a}{$a$} & \node{b}{$b$} & \\
        \nodeconnect[b]{k1}[t]{k2} \nodeconnect[b]{k1}[t]{c}
        \nodeconnect[b]{k2}[t]{b} \nodeconnect[b]{k2}[t]{a}
  \end{tabular}

  \begin{tabular}{p{.5cm}p{.5cm}p{.5cm}p{.5cm}p{.5cm}p{.5cm}}
    \mc{6}{c}{\node{k1}{$\kon$}}\\[1ex]
    \mc{4}{c}{\node{k2}{$\kon$}} & \mc{2}{c}{\node{k3}{$\kon$}}\\[1ex]
    \mc{2}{c}{\node{k4}{$\kon$}} & \mc{2}{c}{\node{k5}{$\kon$}} &
    \node{c1}{$c$} & \node{c2}{$c$}\\[1ex]
    \node{a1}{$a$} & \node{a2}{$a$} & \node{b1}{$b$} & \node{b2}{$b$} & & \\
    \nodeconnect[b]{k1}[t]{k2} \nodeconnect[b]{k1}[t]{k3}
    \nodeconnect[b]{k2}[t]{k4} \nodeconnect[b]{k2}[t]{k5}
    \nodeconnect[b]{k3}[t]{c1} \nodeconnect[b]{k3}[t]{c2}
    \nodeconnect[b]{k4}[t]{a1} \nodeconnect[b]{k4}[t]{a2}
    \nodeconnect[b]{k5}[t]{b1} \nodeconnect[b]{k5}[t]{b2}
  \end{tabular}

  \begin{tabular}{p{.5cm}p{.5cm}p{.5cm}p{.5cm}p{.5cm}p{.5cm}p{.5cm}p{.5cm}p{.5cm}}
    \mc{9}{c}{\node{k1}{$\kon$}}\\[1ex]
    \mc{6}{c}{\node{k2}{$\kon$}} & \mc{3}{c}{\node{k3}{$\kon$}}\\[1ex]
    \mc{3}{c}{\node{k4}{$\kon$}} & \mc{3}{c}{\node{k5}{$\kon$}} &
    \mc{2}{c}{\node{k6}{$\kon$}} & \node{c1}{$c$}\\[1ex]
    \mc{2}{c}{\node{k7}{$\kon$}} & \node{a1}{$a$} &
    \mc{2}{c}{\node{k8}{$\kon$}} & \node{b1}{$b$} &
    \node{c2}{$c$} & \node{c3}{$c$}\\[1ex]
    \node{a2}{$a$} & \node{a3}{$a$} &&
    \node{b2}{$b$} & \node{b3}{$b$} &&&& \\
    \nodeconnect[b]{k1}[t]{k2} \nodeconnect[b]{k1}[t]{k3}
    \nodeconnect[b]{k2}[t]{k4} \nodeconnect[b]{k2}[t]{k5}
    \nodeconnect[b]{k3}[t]{c1} \nodeconnect[b]{k3}[t]{k6}
    \nodeconnect[b]{k4}[t]{a1} \nodeconnect[b]{k4}[t]{k7}
    \nodeconnect[b]{k5}[t]{b1} \nodeconnect[b]{k5}[t]{k8}
    \nodeconnect[b]{k6}[t]{c2} \nodeconnect[b]{k6}[t]{c3}
    \nodeconnect[b]{k7}[t]{a2} \nodeconnect[b]{k7}[t]{a3}
    \nodeconnect[b]{k8}[t]{b2} \nodeconnect[b]{k8}[t]{b3}
  \end{tabular}

    \caption{An Example of $\{a^nb^nc^n\}$}
    \label{fig:anbncnsecond}
  \end{center}
\end{figure}

Transforming the grammar $G$ with the help of the \LIFT\ mapping of
Definition \ref{def:lift} into its derived correspondent $G_D$ produces
a \textsl{regular grammar}. As will be recalled from the remarks after
Definition \ref{def:der-alph}, all symbols from the original alphabet
become constant operators in the derived alphabet. In the presentation
below the coding of type symbols is taken over from the last example in
section \ref{sec:prelim}. It relies upon the bijection between
$S^\ast\times S$ and $\Nat$, where $S$ is a singleton.  Let $\Nat$ be
the set of sorts and let $D(\Sigma)$ be the derived alphabet. The
derived grammar $G_D = \langle D(\Sigma), D(\F),D(S),D(P)\rangle$
contains the following components:
  \[\begin{array}{lclclcl}
    D(\Sigma)_0 & = &\{ a,b,c\} & \hspace*{1cm} & D(\Sigma)_{n,k} &
    = & \{ S \}\\
    D(\Sigma)_2 & = &\{ \kon\} & \hspace*{1cm} & D(\Sigma)_{n} &=
    &\{ \pi \}\\
    &&&&&&\\
    \F_0 &= &\{ S \} & \hspace*{1cm} & \F_3 & = &\{ F \}
  \end{array}\]
  \[
  P = \left\{ \begin{array}{lcl}
      S& \!\ra\! & S(F,a,b,c)\\
      F &\! \ra\!& S(\kon,S(\kon,\pi_1,\pi_2),\pi_3)\,\, |\\
      \mc{3}{l}{\hspace*{.2cm}S(F,S(\kon,\pi_1,S(a)),S(\kon,\pi_2,S(b)),S(\kon,\pi_3,S(c)))}\\
\end{array}\right\}
\]

The context will always distinguish occurrences of the start symbol $S$
from occurrences of the substitution operator $S$.

Sample derivations and two specimens of the generated language
$\Lang(G_D)$ appear in Figures \ref{fig:der1} and \ref{fig:der2}.

\begin{figure}[htp]
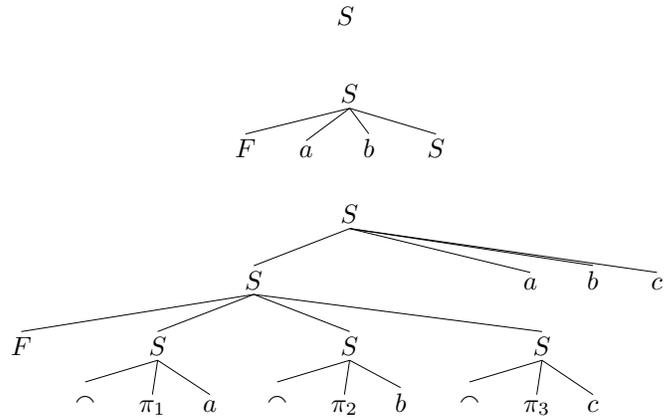

  \begin{center}
    \leavevmode $S$ \vspace*{2em}

    \begin{tabular}{p{.5cm}p{.5cm}p{.5cm}p{.5cm}}
      \mc{4}{c}{\node{k}{$S$}}\\[2ex]
      \node{F}{$F$} & \node{a}{$a$} & \node{b}{$b$} & \node{c}{$S$}\\
      \nodeconnect[b]{k}[t]{F} \nodeconnect[b]{k}[t]{a}
      \nodeconnect[b]{k}[t]{b} \nodeconnect[b]{k}[t]{c}
    \end{tabular}

    \begin{tabular}{p{.5cm}p{.5cm}p{.5cm}p{.5cm}p{.5cm}p{.5cm}p{.5cm}p{.5cm}p{.5cm}p{.5cm}p{.5cm}}
      \mc{11}{c}{\node{k1}{$S$}}\\[3ex]
      \mc{8}{c}{\node{k2}{$S$}} & \node{a1}{$a$} & \node{b1}{$b$} &
      \node{c1}{$c$}\\[3ex]
      \node{F}{$F$} & \mc{3}{c}{\node{k3}{$S$}} &
      \mc{3}{c}{\node{k4}{$S$}} &
      \mc{3}{c}{\node{k5}{$S$}} &\\[2ex]
      & \node{k6}{$\kon$} & \node{p1}{$\pi_1$} & \node{a2}{$a$} &
      \node{k7}{$\kon$} & \node{p2}{$\pi_2$} & \node{b2}{$b$} &
      \node{k8}{$\kon$} & \node{p3}{$\pi_3$} & \node{c2}{$c$} &\\
      \nodeconnect[b]{k1}[t]{k2} \nodeconnect[b]{k1}[t]{a1}
      \nodeconnect[b]{k1}[t]{b1} \nodeconnect[b]{k1}[t]{c1}
      \nodeconnect[b]{k2}[t]{F} \nodeconnect[b]{k2}[t]{k3}
      \nodeconnect[b]{k2}[t]{k4} \nodeconnect[b]{k2}[t]{k5}
      \nodeconnect[b]{k3}[t]{k6} \nodeconnect[b]{k3}[t]{p1}
      \nodeconnect[b]{k3}[t]{a2} \nodeconnect[b]{k4}[t]{k7}
      \nodeconnect[b]{k4}[t]{p2} \nodeconnect[b]{k4}[t]{b2}
      \nodeconnect[b]{k5}[t]{k8} \nodeconnect[b]{k5}[t]{p3}
      \nodeconnect[b]{k5}[t]{c2}
    \end{tabular}
    \caption{Lifted Derivations corresponding to Figure
      \ref{fig:anbncnfirst}.} 
    \label{fig:der1}
  \end{center}
\end{figure}

\begin{figure}[htp]
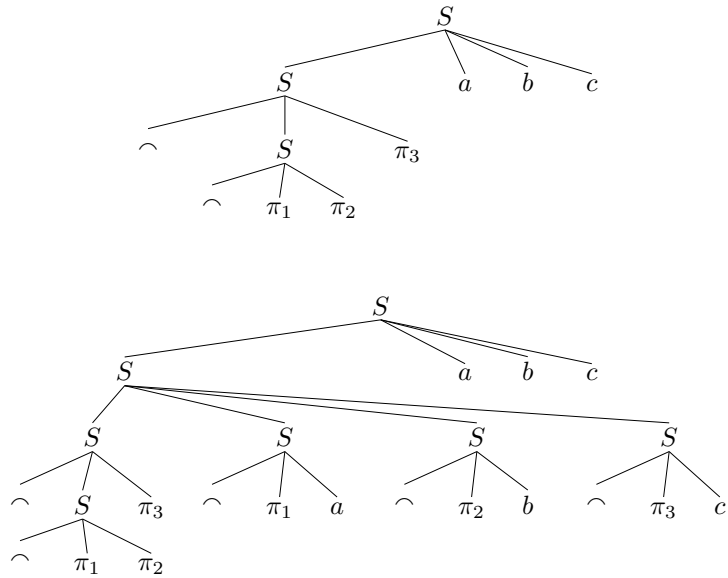

  \begin{center}
    \leavevmode
    \begin{tabular}{p{.5cm}p{.5cm}p{.5cm}p{.5cm}p{.5cm}p{.5cm}p{.5cm}p{.5cm}}
      &&\mc{6}{c}{\node{k1}{$S$}}\\[3ex]
      \mc{5}{c}{\node{k2}{$S$}} & \node{a}{$a$} & \node{b}{$b$} &
      \node{c}{$c$}\\[3ex]
      \node{k3}{$\kon$} & \mc{3}{c}{\node{k4}{$S$}} &
      \node{p3}{$\pi_3$} & &&\\[2ex]
      & \node{k5}{$\kon$} & \node{p1}{$\pi_1$} & \node{p2}{$\pi_2$}
      &&&&\\
      \nodeconnect[b]{k1}[t]{k2} \nodeconnect[b]{k1}[t]{a}
      \nodeconnect[b]{k1}[t]{b} \nodeconnect[b]{k1}[t]{c}
      \nodeconnect[b]{k2}[t]{k3} \nodeconnect[b]{k2}[t]{k4}
      \nodeconnect[b]{k2}[t]{p3} \nodeconnect[b]{k4}[t]{k5}
      \nodeconnect[b]{k4}[t]{p1} \nodeconnect[b]{k4}[t]{p2}
    \end{tabular}

\vspace*{3ex}

\begin{tabular}{p{.5cm}p{.5cm}p{.5cm}p{.5cm}p{.5cm}p{.5cm}p{.5cm}p{.5cm}p{.5cm}p{.5cm}p{.5cm}p{.5cm}}
  \mc{12}{c}{\node{k1}{$S$}}\\[3ex]
  \mc{4}{c}{\node{k6}{$S$}} & &&&
  \node{a}{$a$} & \node{b}{$b$} & \node{c}{$c$}&&\\[3ex]
  \mc{3}{c}{\node{k2}{$S$}} & \mc{3}{c}{\node{k7}{$S$}} &
  \mc{3}{c}{\node{k8}{$S$}} &
  \mc{3}{c}{\node{k9}{$S$}} \\[3ex]
      
  \node{k3}{$\kon$} & \node{k4}{$S$} & \node{p3}{$\pi_3$} &
  \node{k10}{$\kon$} & \node{p4}{$\pi_1$} & \node{a1}{$a$} &
  \node{k11}{$\kon$} & \node{p5}{$\pi_2$} & \node{b1}{$b$} &
  \node{k12}{$\kon$} & \node{p6}{$\pi_3$} & \node{c1}{$c$} \\[2ex]
  \node{k5}{$\kon$} & \node{p1}{$\pi_1$} & \node{p2}{$\pi_2$} & &&&&&&&
  \\
  \nodeconnect[b]{k1}[t]{k6} \nodeconnect[b]{k1}[t]{a}
  \nodeconnect[b]{k1}[t]{b} \nodeconnect[b]{k1}[t]{c}
  \nodeconnect[b]{k6}[t]{k2} \nodeconnect[b]{k6}[t]{k7}
  \nodeconnect[b]{k6}[t]{k8} \nodeconnect[b]{k6}[t]{k9}
  \nodeconnect[b]{k2}[t]{k3} \nodeconnect[b]{k2}[t]{k4}
  \nodeconnect[b]{k2}[t]{p3} \nodeconnect[b]{k4}[t]{k5}
  \nodeconnect[b]{k4}[t]{p1} \nodeconnect[b]{k4}[t]{p2}
  \nodeconnect[b]{k7}[t]{k10} \nodeconnect[b]{k7}[t]{p4}
  \nodeconnect[b]{k7}[t]{a1} \nodeconnect[b]{k8}[t]{k11}
  \nodeconnect[b]{k8}[t]{p5} \nodeconnect[b]{k8}[t]{b1}
  \nodeconnect[b]{k9}[t]{k12} \nodeconnect[b]{k9}[t]{p6}
  \nodeconnect[b]{k9}[t]{c1}
    \end{tabular}
    \caption{Terminal derived trees corresponding to the example in
  Figure \ref{fig:anbncnsecond}.}
    \label{fig:der2}
  \end{center}
\end{figure}
\end{example}

The case illustrated by this example is characteristic of the general
situation. An arbitrary context-free tree grammar $G$ can be mapped into
its derived counterpart $G_D$ with the help of the \LIFT\ 
transformation. The result of this transformation process, $G_D$, is a
regular grammar and therefore specifies a context-free language as the
yield of its generated tree language $\Lang(G_D)$. This follows directly
from the fundamental fact, stated above, that a string language is
context-free if and only if it is the leaf or frontier language of a
regular tree language. The frontiers of $\Lang(G)$ and of $\Lang(G_D)$
are obviously not the same languages. The yield of $\Lang(G_D)$ in
particular consists of strings over the whole alphabet $\Sigma$ extended
by the set of projection symbols. Due to the fact, however, that the
composition of the \LIFT\ operation with the $\beta$ operation is the
identity on the elements of $T(\Sigma,X)$, it is of considerable
interest to know whether this close relationship between elements of
$T(D(\Sigma)$ and of $T(\Sigma,X)$ is preserved by the derivation
process in the context-free grammar and its regular counterpart. Before
we prove a claim about this relationship in the proposition below a
short historical remark appears to be apposite.

The central theorem in \cite{maibaum74} to the effect that every
context-free tree language is the image under the operation $\beta$ of
an effectively constructed regular language is wrong because he
confounded the inside-out with the outside-in derivation mode. In the
course of establishing a fixed-point characterization for context-free
tree grammars in either generation mode \cite{ES77:IO-OI1} point out
this mistake and state as an immediate consequence of the fixed-point
analysis of IO context-free tree grammars within the space of the
power-set tree substitution algebra that each IO context-free tree
language $\Lang$ is the image of a regular tree language $D(\Lang)$
under the unique homomorphism from $\mathfrak{T}(D(\Sigma))$ into
$\mathfrak{DT}(\Sigma,X)$ (see their Cor.4.12). This immediate
consequence is but a restatement of the classical Mezei-Wright result
that the equational subsets of an algebra are the homomorphic images of
recognizable subsets in the initial term algebra. As formulated by
Engelfriet \& Schmidt, their correction of Maibaum's theorem has a
distinctive declarative content whereas the original claim had a clear
operational meaning. It was based on the contention that the individual
derivation steps of a context-free tree grammar and its derived
counterpart correspond to each other. This is the point of the next
lemma.

\smallskip\noindent\textbf{Lemma}\textit{
  Suppose $G = \langle \Sigma, \F, S, P \rangle$ is a context-free tree
  grammar and $\Lang(G)$ its generated tree language. Then there is a
  derived regular tree grammar $G_D = \langle D(\Sigma), D(\F), D(S),
  D(P) \rangle$ such that $\Lang(G)$ is the image of $\Lang(G_D)$ under
  the unique homomorphism from the algebra of tree terms over
  $D(\Sigma\cup\F)$ into the tree substitution algebra over the same
  alphabet. In particular, $t'$ is derived in $G$ from $t$ in $k$ steps,
  i.e.\ $t \Ra t'$ via the productions $\tuple[k]{p}$ in $P$ if and only
  if there are productions $\tuple[k]{p'}$ in $D(P)$ such that
  $\LIFT(t')$ is derived in $G_D$ from $\LIFT(t)$ via the corresponding
  productions.}
\smallskip

\begin{proof}
  The proof is based on the closure of inside-out tree languages under
  tree homomorphisms. The idea of using the \LIFT\ operation for the
  simulation of derivation steps on the derived level can also be found
  in \cite{ES78:IO-OI2}. Let $h_n$ be a family of mappings $h_n:
  \Sigma_n \func T(\Omega,X)$ where $\Sigma$ and $\Omega$ are two ranked
  alphabets. Such a family induces a tree homomorphism $\hat{h}:
  T(\Sigma) \func T(\Omega)$ according to the recursive stipulations:
\[
\begin{array}{rcll}
  \hat{h}(\sigma) & = & h_0(\sigma) & \text{ for } \sigma \in \Sigma_0\\
  \hat{h}(\sigma(\tuple{t})) & = & h_n(\sigma)[\hat{h}(t_1),\dots,
  \hat{h}(t_n)] & \text{ for } \sigma \in \Sigma_n\\
\end{array}
\]
A production $p$ in $P$ can be viewed as determining such a tree
homomorphism $\hat{p}: T(\Sigma\cup\F) \func T(\Sigma\cup\F)$ by
considering the family of mappings $p_n: \Sigma_n \cup \F_n \func
T(\Sigma\cup\F,X_n)$ where $p_n(F) = t$ for $t\in T(\Sigma\cup\F,X_n)$
and $p_n(f) = f(\tuple{x})$ for $f\not= F$ in $\Sigma\cup\F$. By
requiring that $\hat{p}(x_i) = x_i$ the mapping $\hat{p}$ can be
regarded as a $D(\Sigma\cup\F')$-homomorphism from the tree substitution
algebra $\mathfrak{DT}(\Sigma\cup \F,X)$ into itself, where we have set
$\F' := \F \setminus \{F\}$.  By applying the \LIFT-operation to the
tree homomorphism $\hat{p}$ we obtain its simulation $\hat{p}_D:
T(D(\Sigma\cup\F)) \func T(D(\Sigma\cup\F))$ on the derived level:
\[
\begin{array}{rcll}
  p_{D_0}(F) & = & \LIFT_n(p_n(F)) &\\
  p_{D_0}(f) & = & f & \text{ for } f \not= F \text{ in } \Sigma\cup \F\\
  p_{D_0}(\pi_i^n) & = & \pi_i^n &\\
  p_{D_{n+1}}(S_{n,m}) & = & S_{n,m}(\tuple[n+1]{x}) &\\
\end{array}
\]
Observe, that we have treated $D(\Sigma\cup\F)$ as a ranked alphabet. If
we can show that the diagram below commutes the claim in the lemma
follows by induction:
\begin{equation*}
\begin{CD}
  \mathfrak{T}(D(\Sigma\cup\F)) @>{\hat{p}_D}>>
  \mathfrak{T}(D(\Sigma\cup\F)) \\
  @V{\beta}VV @VV{\beta}V\\
  \mathfrak{DT}(\Sigma\cup\F,X) @>>{\hat{p}}>
  \mathfrak{DT}(\Sigma\cup\F,X) \\
\end{CD}
\end{equation*}

The commutability is shown by the succeeding series of equations in
which the decisive step is justified by the identity of
$\beta\circ\LIFT$ on the tree substitution algebra.  Let $f$ be in
$(\Sigma\cup\F)_n$:
\begin{eqnarray*}
  \beta(\hat{p}_D(f)) & = & \beta(\LIFT_n(p_n(f)))\\
  & = & p_n(f)\\
  & = & \hat{p}(f(\tuple{x}))\\
  & = & \hat{p}(\beta(f)).
\end{eqnarray*}
\end{proof}

The preceding result provides an operational handle on the
correspondence between the derivation sequences within the tree
substitution algebra and the derived term algebra. As characterized, the
correspondence is not of much help in finding a solution to the problem
of giving a logical description of the exact computing power needed to
analyze natural language phenomena.  The formal definition of the
$\beta$ transformation, which mediates the correspondence, is of an
appealing perspicuity, but the many structural properties that are
exemplified in the range of this mapping make it difficult to
estimate the computing resources necessary to establish the
correspondence relation between input trees and their values in the
semantic domain of the tree substitution algebra. We know from the
classical result for regular tree languages that monadic second-order
logic is too weak to serve as a logical means to define the range of the
$\beta$ mapping when it is applied to the space of regular tree
languages.  What does not seem to be excluded is the possibility of
solving our logical characterization problem by defining the range of
context-free tree languages \textsl{within} the domain of the regular
languages. In this way, we would rest on firm ground and would take a
glimpse into unknown territory by using the same logical instruments
that helped us to survey our "recognizable" homeland.

\section{Logical Characterization}

Extending the characterization of grammatical properties by monadic
second-order logic has been our main motivation. For tree languages the
central result is the following: a tree language is definable in monadic
second-order logic if and only if it is a regular language. As is well
known and as we will indicate below, a similar characterization holds
for regular and context-free string languages. The examples and analyses
of English syntactic phenomena that are presented in \cite{roge:stud94}
make it abundantly clear that monadic second-order logic can be used as
a flexible and powerful specification language for a wide range of
theoretical constructs that form the core of one of the leading
linguistic models.  Therefore, the logical definability of structural
properties that are coextensive with the empirically testified
structural variation of natural languages, would be a useful and
entirely grammar independent characterization of the notion of a possible
human language.

It follows from the cross-serial dependencies in Swiss German and
related phenomena in other languages that monadic second-order logic, at
least in the form that subtends the characterization results just
mentioned, is not expressive enough to allow for a logical solution of
the main problem of Universal Grammar: determining the range in
cross-language variation. In many minds, this expressive weakness alone
disqualifies monadic second-order logic from consideration in
metatheoretic studies on the logical foundations of linguistics.
Employing the results of the preceding section, we will sketch a way out
of this quandary, inspired by a familiar logical technique of talking in
one structure about another. To what extent one can, based on this
technique, simulate transformations of trees is not yet fully
understood. We shall sketch some further lines of research in the
concluding remarks.

The major aim of descriptive complexity theory consists in classifying
properties and problems according to the logical complexity of the
formulas in which they are expressible. One of the first results in this
area of research is the definability of the classes of regular string
and tree languages by means of monadic second-order logic. The use of
this logic is of particular interest since it is powerful enough to
express structural properties of practical relevance in the empirical
sciences and it remains, in spite of its practical importance,
efficiently solvable. As a preparation for our logical description of
the $\beta$ transformation we shall recall some concepts needed for
expressing the logical characterization of the class of regular
languages.

For the purpose of logical definability we regard strings over a finite
alphabet as model-theoretic structures of a certain kind.

\begin{definition}
  Let $\Sigma$ be an alphabet and let $\tau(\Sigma)$ be the vocabulary
  \mbox{$\{<\}$} $\null\cup\{P_a\,|\,a\in\Sigma\}$, where $<$ is binary
  and the $P_a$ are monadic. A \textbf{word model} for $u\in\Sigma^\ast$
  is a structure of the form
    \[ \M = (B,<,P_a)\]
    where $|B| = length(u)$, $<$ is an ordering of $B$ and the $P_a$
    correspond to positions in $u$ carrying the label $a$:
    \[ P_a= \{ b\in B \, |\, label(b)=a\}.\]

\end{definition}
  
The corresponding monadic second-order logic $\MSOL(\Sigma)$ has
\emph{node variables} $x,y,z,\dots$ and \emph{set variables}
$X,Y,Z,\dots$ ranging over subsets of the domain of discourse $B$.
There are four kinds of atomic formulas $x=y$, $x<y$, $P_a(x)$ and $x\in
X$. The formulas of the language are constructed from the atomic
formulas in the expected way by combining them by means of the usual
propositional connectives and the existential and universal quantifiers.
Observe that not only individual node variables, but also node-set
variables may be bound by quantifiers. For a closed formula $A$ in this
language and a word model $\M$, we write $\M\models A$ if $A$ is true in
$M$. The set of models that make a closed formula $A$ true is denoted by
$Mod(A)$.  A theorem by Elgot talks about the close relationship between
model classes definable by closed formulas of $\MSOL(\Sigma)$ and
regular string languages.

\begin{theorem}[Elgot] $\Lang$ is regular over the alphabet $\Sigma$ iff
  $\Lang$ is definable in monadic second-order logic over the vocabulary
  $\tau(\Sigma)$. More succinctly:
  \[
    \Lang \in \Lang_3 \text{ iff }
    \exists A \in \MSOL(\Sigma)\text{ such that }
    \Lang = Mod(A)= \{ u\in \Sigma^\ast \, |\, u\models A \}.
  \]
\end{theorem}

For the statement of this classical result for tree languages we need an
appropriate notion of model-theoretic structure. If $\Sigma$ is a finite
ranked alphabet, a tree in $T(\Sigma)$ is coded by a model of the
following form, which is called a \emph{labeled tree model}.

\begin{definition}
  Let $\Sigma$ be a finite ranked alphabet and let $\tau(\Sigma)$ be the
  vocabulary $\{<_i\}_{i\leq n}\cup\{P_a\,|\,a\in\Sigma\}$, where each
  $<_i$ is binary and each $P_a$ is a monadic predicate. A
  \textbf{labeled tree model} for $t\in T_\Sigma$ is a structure of the
  form
    \[ M= (B,<_i,P_a)_{i \leq n, a\in \Sigma}\]
    where $B$ is in one-one correspondence with $dom(t)$, the tree
    domain of $t$, $<_i$ is the $i$-th successor function, $n =\null$
    the maximal $m$ such that $\Sigma_m \not= \emptyset$ and the $P_a$
    correspond to positions with label $a$:
    \[ P_a= \{ b\in B \, |\, label(b)=a\}.\]
\end{definition}

The notion of a labeled tree is built upon a system of nodes or tree
addresses which are strings in $\Nat^\ast$ and which satisfy two closure
properties. More precisely, the \emph{domain} of a finite ordered tree
with at most $n$ outgoing branches is a finite subset $D\subseteq
\Nat_n^\ast$, where $\Nat_n = \{ 0,\dots,n-1\}$, such that the following
conditions hold:
\begin{enumerate}
  \renewcommand{\labelenumi}{(\roman{enumi})}
\item If $uv\in D$ ($u,v \in \Nat_n^\ast$) then $u\in D$.
\item If $ui\in D$ and $j < i$ then $uj \in D$ ($u\in \Nat_n^\ast$,
  $i\in \Nat_n$). 
\end{enumerate}

The appropriate \textsl{monadic second-order logic of $n$ successors}
$\MSOL_n(\Sigma)$ has \emph{node variables} $x,y,z,\dots$ and \emph{set
  variables} $X,Y,Z,\dots$, ranging over subsets of the tree domain.
There are again four kinds of atomic formulas $x=y$, $x<_i y$, $P_a(x)$
and $x\in X$, where we have presented the successor function in
relational form. Arbitrary formulas are generated from atomic formulas
by propositional connectives and the quantifiers binding both kinds of
variables. If $A$ is an $\MSOL_n(\Sigma)$ sentence, i.e.\ a formula
without free variables, and $\M$ a tree model, we write $\M\models A$ to
indicate that $A$ is true in $\M$. If $Mod(A) = \{ \M \, | \, \M \models
A \}$, we call this set of trees \emph{definable} in $\MSOL_n(\Sigma)$.
The next theorem states the extension of Elgot's result to the yield or
frontier of finite ranked ordered trees.

\begin{theorem}[Doner, Thatcher--Wright]
  Let $\Sigma$ be a ranked alphabet.
\begin{enumerate}
  \renewcommand{\labelenumi}{\alph{enumi})}
  \renewcommand{\theenumi}{\alph{enumi}}
\item $\Lang\subseteq\Sigma^\ast$ is \textbf{context-free} iff $\Lang$
  is the yield of a set of tree models definable in monadic second-order
  logic of $n$ successors over the vocabulary $\tau(\Sigma)$.
  \[
  \Lang \in \Lang_2 \text{ iff } \exists A \in \MSOL_n(\Sigma) \text{
    such that } \Lang = yield(Mod(A))
  \]
  where $Mod(A) = \{t\in T_\Sigma \, |\, t\models A \}.$
\item A set of trees is generated by a regular tree grammar $G$ iff
  there is a sentence $A$ in $\MSOL_n(\Sigma)$ such that
    \[ Mod(A) = \Lang(G). \]
  \end{enumerate}
\end{theorem}

Given the intimate relationship between regular tree languages and
context-free string languages on the one hand and monadic second-order
logic on the other, there is no hope to address the question as to how
to account for context-sensitive formalisms with the definitional
resources of this logic directly. An alternative approach is suggested
by the characterization of the context-free tree languages as the class
of languages that is the homomorphic image of the class of regular tree
languages over a related alphabet. This characterization says that the
context-free tree languages constitute the class of all tree families
$Tr(\Lang)$ where $\Lang$ is a subset of $T(D(\Sigma))$ and $Tr$ is a
function from explicit trees over $D(\Sigma)$ to trees over $\Sigma$.

The $\beta$ transformation that sends regular tree languages to their
context-free counterparts is a particular instance of a relation that
mediates be\-tween finite trees over different alphabets. If $\Sigma$ and
$\Omega$ are two alphabets, then a binary relation $Tr \subseteq
T(\Sigma) \times T(\Omega)$ is generally called a \emph{tree
  transduction}. Tree transductions can be interpreted as transforming a
tree over one alphabet into a tree over another alphabet. Of special
interest are those tree transductions that can be specified in an
effective way and among these, the top-down and bottom-up transductions,
performed by finite automata that process a tree from the root down or
from the leaves up, have been thoroughly investigated since the early
70's. It has been shown that these types of transductions do not
preserve regularity, but that their range is a proper subset of the
context-free tree languages when their domain is restricted to the
regular tree languages. Since the range of the $\beta$ transformation is
the full range of context-free tree languages under the same domain
restriction its action cannot be simulated by a top-down or a bottom-up
tree transduction.

As the reader will recall, the transformations performed by the $\beta$
function interpret the operators $S$ and $\pi$ as substitution and
projection, respectively. This behaviour is of such a formal simplicity
that an effective description of its structural pattern should be
possible, even though a simulation by a classical (top-down or
bottom-up) tree automaton is excluded according to the result just
mentioned. Trees in $T(D(\Sigma))$ and trees in $T(\Sigma)$ are really
different ways of looking at one and the same thing. Therefore, it would
further the project of an effective description of tree classes in the
substitution algebra if there was an effective syntactic map from the
vocabulary $D(\Sigma)$ into the vocabulary $\Sigma$, because we know
already that monadic second-order logic provides an effective medium to
define the regular tree classes in the initial $D(\Sigma)$-algebra.

There is an established technique in model theory that enables one to
show that in certain circumstances the choice of language is arbitrary.
The central notion employed in this technique is that of a syntactic
interpretation. The interesting feature of \textsl{syntactic
  interpretation} is that it allows one to talk in one structure about
another, as we will now attempt to show.  Suppose we have two relational
signatures $\Sigma$ and $\Omega$ with no constant or function symbols
and, in addition to a \textsl{domain formula} $A_\Omega(x)$, for each
$n$-ary $R\in \Omega$ a \textsl{defining formula} $A_R(\tuple{x})$,
where these formulas are built from the symbols in $\Sigma$. Given a
$\Sigma$-structure $\M$ in which the domain formula $A_\Omega(x)$
defines the universe of discourse of an $\Omega$-structure $\M'$, it can
then be shown that for every $\Omega$-formula $B$ one can find an
$\Sigma$-formula $B'$ such that
\[ \M \models B'\,[g] \text{ iff } \M' \models B\, [g] \]
where $g$ is a variable assignment in $\M'$.

Note, that from a different perspective the domain formula $A_\Omega(x)$
can be viewed as establishing a binary relation between $\Sigma$- and
$\Omega$-structures in the following way. We let $A_{\Omega}$ single out
those $\Sigma$-structures where its satisfaction set is nonempty:
\[ \M' = \{ a \, | \, \M \models A_\Omega[a] \} \not= \emptyset. \]
On $\M'$ an $\Omega$-structure is easily specified by the following
stipulation for each $R\in \Omega$:
\begin{align*}
  R_{\M'} := \{ (\tuple{a}) \, | \, & \M \models A_R[\tuple{a}] \text{ \& }\\
  & \M \models A_\Omega[a_i] \text{ for } i = 1,\dots,n \}
\end{align*}
(In case $\Omega$ contains any functional relations appropriate closure
conditions have to be required in addition to the non-emptiness of
$\M'$.) As defined, $A_\Omega$ is nothing but a binary relation
$A_\Omega \subseteq \mathbf{M}_\Sigma \times \mathbf{M}_\Omega$ where
$\mathbf{M}_\Sigma$ denotes the class of $\Sigma$-structures and
$\mathbf{M}_\Omega$ the class of $\Omega$-structures: Two structures
$\M$ and $\M'$ stand in the relation $A_\Omega$ if $\M$ is a
$\Sigma$-structure in which we can talk about an internal
$\Omega$-structure $\M'$ that is carved out by the satisfaction of
$A_{\Omega}$.

The above analysis involves the obvious suggestion to provide an account
of tree transductions within the model-theoretic framework of syntactic
interpretation. Once labeled trees are looked at as specific
model-theoretic structures a tree transduction $Tr$ becomes a relation
between such structures in perfect analogy to the case of relational
structures connected by a set of defining and domain formulas.

The first to see the potential of applying the method of syntactic
interpretation to the analysis of relations between graph-like
structures was B. Courcelle. He introduces for this purpose the notion
of a monadic second-order definable graph transduction. The switch from
first-order logic to monadic second-order logic is caused by the same
reasons that were given above for the adoption of monadic second-order
logic in the case of labeled ordered trees.  Apart from this recourse to
higher-order logic Courcelle's definition is patterned upon the
classical format provided by the method of syntactic interpretation. Let
us record here its adaptation to tree structures.

\begin{definition}
  Let $\Sigma$ and $\Omega$ be tow finite ranked alphabets. A
  \textbf{monadic second-order definable tree transduction} is defined
  as follows. We fix a syntactic interpretation $I = ( A, A_\Omega(x),
  A_{<_i}(x,y), A_{P_a}(x))_{i\leq n, a\in \Omega}$ consisting of a
  tuple of monadic second-order formulas over the signature $\Sigma$.
  These formulas are intended to define a tree model $t'$ in $T(\Omega)$
  from a tree model $t$ in $T(\Sigma)$. The closed formula $A$ is to
  define the domain of the transduction, the formula $A_\Omega(x)$ is to
  define the tree domain of $t'$ and the formulas $A_{<_i}(x,y)$ and
  $A_{P_a}(x)$ are to define the successor relations and the
  distribution of node labels on the tree domain $dom(t')$. $n$
  indicates the maximum arity of a symbol in $\Omega$.
  
  A tree model $t'$ in $T(\Omega)$ with tree domain $D$ is defined in
  $t\in T(\Sigma)$ by $I$ if
\begin{enumerate}
  \renewcommand{\labelenumi}{\alph{enumi})}
  \renewcommand{\labelenumii}{(\roman{enumii})}
  \renewcommand{\theenumi}{\alph{enumi}}
  \renewcommand{\theenumii}{\roman{enumii}}
\item
  \begin{enumerate}
  \item $t\models A$
  \item $D = \{ u \, | \, t \models A_\Omega\,[u] \}$
  \item $<_{i_{t'}} = \{ (u,v) \in D^2 \, | \, t \models A_{<_i}\,[u,v]
    \}$
  \item $P_{a_{t'}} = \{ u \in D \, | \, t \models A_{P_a}\, [u] \}$
  \end{enumerate}
\item $D$ satisfies the conditions of a tree domain with $<_{i_{t'}}$
  its successor relations.
\end{enumerate}
Indicating this last (functional) relation by $def_I(t) = t'$ we can
finally say what is denoted by the monadic second-order tree
transduction defined by I:
\[
def_I = \{ (t,t')\, | \, def_I(t) = t' \}.
\]
\end{definition}

As defined, a definable tree transduction is indeed a tree transduction
in the sense of the abstract transformation relation $Tr\subseteq
T(\Sigma)\times T(\Omega)$ between tree domains that was introduced
above. What is more, definable transductions possess the sort of
computational properties that we asked of an effective syntactic
mapping. The syntactic interpretation is a finitary object, consisting
of a tuple of formulas, and all the conditions in the definition are
expressible by formulas of the monadic second-order language, which are
decidable in general and even testable in linear time on given tree
models. In other words, the co-domain of a definable transduction
$def_I$ may not be definable itself, i.e.\ there may be no monadic
second-order formula $A'$ such that $\{ t \, | \, t \models A' \} = \{
t' \, | \, def_I(t) = t' \text{ for some } t \text{ s.t.\ } t\models A
\}$ where $A$ defines the domain of $def_I$. The second-order theory of
the co-domain of a definable transduction $def_I$, on the other hand, is
decidable and we can call it even definable if its domain formula is
admitted as a restricting parameter.

Definable tree transductions would thus provide a solution to the
logical characterization problem if the type of tree transformation
performed by the $\beta$ mapping could be given a logical definition via
a syntactic interpretation scheme. This is fortunately the case as
stated in the next theorem.

\begin{theorem}
  The $\beta$ transformation of a regular tree language is a monadic
  second-order definable tree translation.
\end{theorem}

\begin{proof}[Proof Sketch]
  The detailed verification of the theorem's claim is left for another
  occasion. Let us point out that it can be reduced to the definability
  of the yield function of graph grammars with neighbourhood controlled
  embedding and dynamic edge relabeling (edNCE grammars). Productions in
  this type of grammar are of the form $X \func (D,C)$ where $X$ is an
  operative or functional node label, $D$ is a graph and $C$ is a set of
  connection instructions. A derivation step licensed by such a
  production consists in replacing an $X$-labeled node by $D$ and
  connecting $D$ to the neighbourhood of $X$ according to the
  specifications in $C$. As will be recalled, a much more specific form
  of local node replacement played a central role in macro-like
  productions of context-free tree grammars. It is straightforward to
  associate a derivation tree with a derivation in a edNCE grammar. A
  derivation is a labeled ordered tree in which the nodes are labeled
  with productions. If a node has the label $X \func (D,C)$ its
  daughters are labeled by a sequence of production rules whose
  left-hand sides exhaust the operative nodes in the replacing graph
  $D$. The \emph{yield} function executes the substitutions of the
  right-hand sides of the production rules in a bottom-up regime,
  observing along the way the connecting restrictions. The yield of the
  derivation tree returns as result the graph that is generated by a
  corresponding derivation of the edNCE grammar.
  
  The fact that the yield function of an edNCE grammar is definable in
  monadic second-order logic was proved by van Oostrom and presented in
  a generalized form in \cite{cour:them92}. The trees in a regular tree
  grammar over a derived alphabet can be read as a special form of
  derivation trees in the sense of edNCE graph grammars. In these
  special trees, the replacements of the functional symbols have already
  been carried out, but the connecting instructions which are coded in
  the explicit substitution and projection labels have to be executed by
  the $\beta$ transformations.  Thus, the $\beta$ transformation turns
  out to be a special case of the yield function, which means that the
  $\beta$ transformation can be simulated by a definable transduction. A
  detailed proof of the claim that the case of context-free tree
  languages is indeed covered by the context-free neighbourhood
  controlled graph languages will appear elsewhere.
\end{proof}

\section{Concluding Remarks}

In the present paper, we have tried to combine the methods developed in
the tradition of algebraic language theory with the techniques of model
theory. While there is a perfect match between these two methodological
frameworks in the case of regular or, equivalently, taking the
automata-theoretic view, recognizable families of trees, this state of
harmony is seriously disturbed once one considers the far richer family
of context-free sets of trees. We had no choice but to take these richer
structures into account, since a series of well documented non-local
dependencies in natural languages fall outside the range of
configurational properties that can be accommodated within the realm of
regular trees. The grammar independent approach to characterizing
syntactic concepts has been so rewarding, on the other hand, that it
seemed worthwhile to make a serious attempt at providing the formal
background to extend the tools of the logical analysis beyond the
harmonious domain of strictly local dependencies.

As was emphasized several times above, it may well be possible to
strengthen the language of monadic second-order logic without losing its
algorithmic tractability. Whether such a strengthening would create a
framework for a direct application of monadic second-order logic's
definitional resources to the domain of context-free tree languages
remains a challenging open problem.

Deterred by the apparent obstacles to devising a manageable logical
specification language for non-local dependencies we opted for the timid
alternative, which consists in forming a picture of the outer world in
terms of the conceptual network that fits so smoothly the regular homeland.
This description of our approach refers both to the algebraic point of
view according to which we evaluate (uniquely) the regular term
expressions in the semantic domain of context-free tree languages and to
the logical point of view according to which we consider the
recognizable forest of trees as relational structures in the
model-theoretic sense and employ their language to talk about the
foreign country where non-local dependencies roam.

It is notoriously difficult to tell to what an extent the use of our own
vernacular gives a distorted picture of a foreign country.  The
complexity of the clauses that enter into an explicit definition of the
evaluation transduction would give an idea of how faithful our language
reflects the situation in an area where it has not been used as a proven
means of communication. We had to postpone the task of spelling out the
translation scheme involved in such an evaluating transduction but the
reader will be able to measure the distance between a syntactic
configuration under its intended representation and its "regular"
counterpart after a perusal of H.-P.~Kolb's contribution to the present
volume.

There is a last disclaimer that needs to be made. The term definable
transduction imports the misleading notion of an operational
transformation of structures. This misinterpretation would bestow a
power upon the device of syntactic interpretation that it definitely
does not have. The issue of whether certain linguistic phenomena are
more easily described in a derivational rather than a representational
model should not be confused with the issue of whether it is possible to
give a direct logical definition of context-free tree structures or only
an indirect definition via a definable transduction. The fact that we
have opted for an indirect solution to our problem testifies only to our
lack of ingenuity and not to our allegiance to the derivationists'
cause.

\nocite{arno:unth76}
\nocite{bena:stru68}
\nocite{bloo:iter93}
\nocite{gall:n-ra84}
\nocite{gecseg&steinby:84}
\nocite{hoeh:supe72}
\nocite{mcke:alge87}
\nocite{rabin69}

\newpage
\bibliographystyle{dcu} 

\end{document}